\documentclass[10pt,twocolumn,letterpaper]{article}

\usepackage{cvpr}              

\definecolor{cvprblue}{rgb}{0.21,0.49,0.74}
\usepackage[pagebackref,breaklinks,colorlinks,allcolors=cvprblue]{hyperref}

\usepackage{lipsum}
\usepackage{svg}
\usepackage{hyperref}
\usepackage{url}
\usepackage{booktabs}
\usepackage{amssymb}
\usepackage{pifont}
\usepackage{multirow}
\usepackage{multicol}
\usepackage{nicefrac}
\usepackage{bm}
\usepackage{graphicx}
\usepackage{colortbl}
\usepackage{amsmath, amssymb, amsthm}
\usepackage{booktabs}
\usepackage{siunitx}
\usepackage{graphicx}
\usepackage{xcolor}
\usepackage{multicol}
\usepackage{colortbl}
\usepackage{tabularx}
\usepackage{makecell}
\usepackage{multirow}
\usepackage{diagbox}
\usepackage{wrapfig}
\usepackage{enumitem}
\usepackage{algorithm, algorithmic}
\usepackage[normalem]{ulem}
\usepackage{tgcursor}
\usepackage{bbm}
\usepackage[utf8]{inputenc}
\usepackage{amsmath}
\usepackage{amssymb}
\usepackage{url}
\usepackage{enumitem}
\usepackage{caption}
\usepackage{tocloft}
\usepackage{pifont}

\usepackage{svg}
\usepackage{amsthm}
\usepackage{tablefootnote}
\usepackage{threeparttable}
\usepackage{float}

\def\paperID{***} 
\def\confName{CVPR}
\def\confYear{2026}

\title{DisCa: Accelerating Video Diffusion Transformers with Distillation-Compatible Learnable Feature Caching}

\def\spaces{~~~~~}
\def\lessspaces{~~~}
\author{Chang Zou$^{1,2}$\thanks{shenyizou@gmail.com},  \lessspaces{}  
Changlin Li$^2$,            \lessspaces{} 
Yang Li$^2$,                \lessspaces{}
Patrol Li$^2$,              \lessspaces{}
Jianbing Wu$^2$,            \lessspaces{}
Xiao He$^{2,3}$,            \\
Songtao Liu$^2$,            \spaces{}
Zhao Zhong$^2$,            \spaces{}
Kailin Huang$^1$,            \spaces{}
Linfeng Zhang$^1$\thanks{Corresponding Author.} \\\\
$^1$Shanghai Jiao Tong University ~~
$^2$Tencent Hunyuan ~~
$^3$Xidian University\\ 
}

\begin{document}
\maketitle
\begin{abstract}
While diffusion models have achieved great success in the field of video generation, this progress is accompanied by a rapidly escalating computational burden.
Among the existing acceleration methods, Feature Caching is popular due to its training-free property and considerable speedup performance,
but it inevitably faces semantic and detail drop with further compression. Another widely adopted method, training-aware step-distillation, though successful in image generation, also faces drastic degradation in video generation with a few steps. 
Furthermore, the quality loss becomes more severe when simply applying training-free feature caching to the step-distilled models, due to the sparser sampling steps. 
This paper novelly introduces a distillation-compatible learnable feature caching mechanism for the first time. We employ a lightweight learnable neural predictor instead of traditional training-free heuristics for diffusion models, enabling a more accurate capture of the high-dimensional feature evolution process. Furthermore, we explore the challenges of highly compressed distillation on large-scale video models and propose a conservative Restricted MeanFlow approach to achieve more stable and lossless distillation. 
By undertaking these initiatives, we further push the acceleration boundaries to $11.8\times$ while preserving generation quality. Extensive experiments demonstrate the effectiveness of our method. \\ 
Code: \textbf{\href{https://github.com/Tencent-Hunyuan/DisCa}{{\textcolor{cyan}{https://github.com/Tencent-Hunyuan/DisCa}}}}
\end{abstract}
\section{Introduction}
\label{sec:intro}
\vspace{-3mm}
In recent years, Diffusion Models (DMs)~\cite{sohl2015deep,ho2020DDPM, DM} have achieved remarkable success in the generative domain, including but not limited to modalities such as image~\cite{rombach2022SD,flux2024}, video~\cite{blattmann2023SVD,open_sora_plan,openai2024sora,wan_wan_2025,hunyuanvideo2025}, audio~\cite{Eskimez2024E2TE,chen-etal-2024-f5tts}, and text~\cite{Nie2025LargeLD, Zhu2025LLaDA1V} generation. To further enhance generation quality, the scale of DMs has been rapidly increasing~\cite{wu2025qwenimagetechnicalreport,li_hunyuan-dit_2024, HunyuanImage-2.1, yang2025cogvideox}, accompanied by a continuous rise in their computational load, making the deployment costs of these models prohibitively high.

To address this issue, numerous acceleration methods have been proposed \cite{yuan2024ditfastattn,zou2024accelerating, Lu2025SimplifyingSA, Yin2024FromSB, Zhang2025AccVideoAV}, primarily focusing on Sampling Timestep Reduction and Denoising Network Acceleration. Among these, training-aware step distillation \cite{Geng2025MeanFF,yin2024stepdistillation, Yin2023OneStepDW} and training-free feature caching methods \cite{ma2024deepcache,liu2024timestep, Geng2024OmniCacheAU} have demonstrated superior performance and have been widely adopted, respectively.

\begin{figure}
\centering\includegraphics[width=\linewidth]{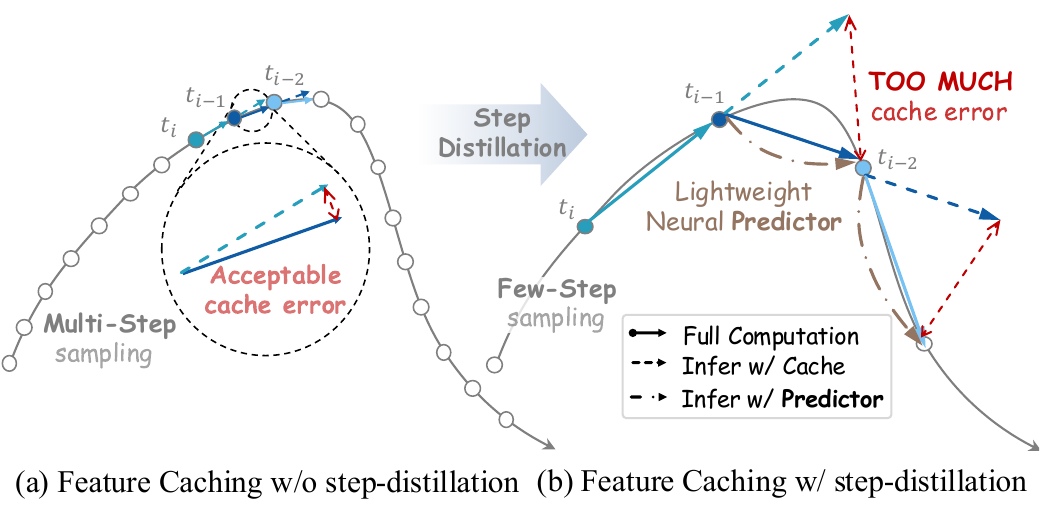}
\vspace{-7mm}
\caption{\textbf{Feature Caching on the diffusion sampling process w/ and w/o step-distillation.} 
(a) Adjacent timesteps are similar under the undistilled models, allowing traditional caching with simple reuse/interpolation. (b) Significant inter-step differences cause traditional caching to fail; the proposed learnable predictor captures the high-dimensional feature evolution successfully.
}
\vspace{-7mm}

\label{fig:Heading}
\end{figure}

Recently, MeanFlow~\cite{Geng2025MeanFF}, as one of the representative methods in step-distillation, builds upon prior research~\cite{Lu2025SimplifyingSA, Frans2024OneSD} with a breakthrough by shifting the sampling objective of DMs from instantaneous velocity to average velocity. 

However, when applied to distill \textit{highly complex large video generation models}~\cite{kong2024hunyuanvideo,wan_wan_2025}, the numerical errors led by its differentiation operations (i.e., Jacobian-vector product (JVP) operations), coupled with MeanFlow's inherently too aggressive design for one-step distillation, can result in catastrophic training divergence and generative artifacts.

Traditional training-free feature caching methods primarily accelerate computation by caching and reusing all~\cite{liu2024timestep, Lv2024FasterCacheTV, li2023FasterDiffusion} or partial~\cite{ma2024l2c,zou2024accelerating,ma2024deepcache,kahatapitiyaAdaptiveCachingFaster2024} previously computed features, thereby skipping relatively non-critical computations. However, since the features at adjacent timesteps in DMs are inherently different, this approach inevitably introduces cache errors. Recent work, TaylorSeer~\cite{TaylorSeer2025}, observed the continuous trajectories of the diffusion sampling procedure. By leveraging multi-step historical features via Taylor series expansion, it significantly reduced cache errors, pushing the acceleration limits of feature caching methods. However, such solutions, which involve manually constructing prediction functions based on specific prior assumptions, are inherently limited in their ability to fully capture the evolutionary trends of high-dimensional features in diffusion models. Consequently, these traditional training-free cache-based methods still exhibit noticeable loss of high-frequency and semantic details.

This paper begins with an intuitive classical phenomenon: the inherent loss of detail in traditional feature caching methods becomes more severe in step-distilled models. As shown in Figure \ref{fig:Heading}(a), the sampling trajectory of diffusion models from noise to data is inherently highly continuous when using a large number of inference steps, which is precisely where traditional training-free feature caching proves effective. However, after the model undergoes step-distillation, the sampling points on the noise-data pair trajectory become sparse. Consequently, the velocity predictions generated by the step-distilled model exhibit significant discrepancies as in Figure \ref{fig:Heading}(b). This makes it challenging to utilize previous features to guide and predict the model's subsequent velocity evolution, as it can no longer be achieved through simple linear or higher-order elementary functions in traditional feature caching methods.

Therefore, we innovatively introduce \textit{lightweight learnable neural network predictors}, rather than handcrafted prediction formulas, in such challenging prediction scenarios led by step-distillation. Through this powerful data-driven approach, the lightweight neural network predictor can effectively capture the evolutionary trends of high-dimensional features during diffusion sampling, thereby achieving more accurate predictions of these features.

Additionally, \textit{improving the stability of step-distillation} also contributes to enhancing the final generation quality. We observed that when applied to large-scale, complex video generation models, MeanFlow cannot achieve the goal of inference in just a few steps while preserving quality. In such scenarios with high-quality requirements, the original few-step or even one-step distillation methods become overly aggressive, negatively impacting the distillation process. Therefore, \textit{Restricted MeanFlow} is designed by limiting the sampling range of average velocities during the MeanFlow distillation process and pruning cases with excessively high compression ratios in the MeanFlow training, thereby enabling more stable distillation results.

Through dual improvements to the feature caching and step-distillation methods, traditional training-free and training-aware approaches are made compatible, further pushing the boundaries of acceleration limits for large-scale video generation models under high-quality generation.

In summary, our main contributions are as follows:
\begin{itemize}
\item \textbf{Lightweight Neural Predictor for Caching}: We propose a \textit{lightweight learnable neural predictor}, enabling further acceleration in step-distilled models with much more difficult sampling trajectories. To the best of our knowledge, the proposed \textit{Dis}tillation-Compatible Learnable Feature \textit{Ca}ching (DisCa) is the first to suggest a `learnable feature caching with distillation' solution.
\item \textbf{Restricted MeanFlow}: We discuss the challenges for highly compressed distillation on large-scale video models. Through pruning the highly compressed scenarios in MeanFlow training, we provide a conservative scheme, Restricted MeanFlow, for a more stable video generation.
\item \textbf{State-of-the-Art Performance}: The proposed DisCa, with Restricted MeanFlow, has highly outperformed the previous methods with an almost lossless acceleration of $11.8\times$. By enabling the training-free and training-aware acceleration methods complementing each other, DisCa provides a new pathway for efficient generation.
\end{itemize}
\vspace{-2mm}

\section{Related Works}
\vspace{-2mm}
\label{sec:related_works}

Diffusion Models (DMs)~\cite{sohl2015deep,ho2020DDPM} have increasingly demonstrated superior performance across many modalities and tasks, with particularly outstanding quality in image and video generation tasks. In recent years, the model architecture of DMs has shown a trend of evolving from the traditional U-Net structure ~\cite{ronneberger2015unet} to Diffusion Transformers (DiTs) ~\cite{peebles2023dit}, primarily due to the excellent scalability of DiTs. While increasing the scale of the models enhances the quality of the generated content, this improvement comes at the cost of a substantially higher computational load for the denoising models. This issue is particularly severe in current state-of-the-art (SOTA) diffusion models~\cite{chen2023pixartalpha,chen2024pixartsigma,opensora,yang2025cogvideox,li_hunyuan-dit_2024, kong2024hunyuanvideo, wan_wan_2025, HunyuanImage-2.1}, which require dozens of iterative steps for the denoising process. Consequently, the demand for accelerating diffusion models has become increasingly urgent. Currently, acceleration techniques for diffusion models can mainly be categorized into \textit{Sampling Timestep Reduction} and \textit{Denoising Network Acceleration}.
\vspace{-1mm}
\subsection{Sampling Timestep Reduction}
\vspace{-2mm}
A straightforward and intuitive approach to acceleration is to \textit{reduce sampling steps while preserving output quality}. DDIM~\cite{songDDIM} introduced a deterministic sampling framework that maintained generation fidelity with fewer denoising iterations. Subsequent advancements, such as the DPM-Solver series~\cite{lu2022dpm,lu2022dpm++,zheng2023dpmsolvervF}, enhanced this direction through high-order numerical ODE solvers. Flow matching~\cite{Lipman2022FlowMF, refitiedflow} further generalizes the diffusion process into a transformation between noise and data points, defining the evolution path of probability distributions via a deterministic instant velocity field, improving sampling efficiency.
\textit{Step distillation methods}~\cite{salimans2022progressive,meng2022on,Yin2023OneStepDW,Salimans2024MultistepDO,yin2024stepdistillation} and \textit{Consistency models}~\cite{song2023consistency, Luo2023LatentCM, Lv2025DCMDC} consolidate multiple denoising operations into a reduced number of steps or even a single sampling pass. The Shortcut model~\cite{Frans2024OneSD} cleverly combines the two by enabling the model to learn the distance between two points on the sampling trajectory. Building on this, Meanflow~\cite{Geng2025MeanFF} innovatively proposes modeling the mean velocity field instead of the instantaneous velocity field, demonstrating outstanding performance in the image generation task and even showing considerable results with one-step generation.
However, translating this success to video generation, which is characterized by significantly higher complexity, reveals limitations. Noticeable declines in quality are often exhibited there, even under conservative multi-step or few-step distillation scenarios. Consequently, there is a strong necessity for \textit{a distillation scheme tailored to the demands of large models and high-quality video generation}.
\vspace{-1mm}
\subsection{Denoising Network Acceleration}
\vspace{-2mm}
\textit{Improving the single-pass computational efficiency} of the denoising network is another effective direction for enhancing the computational efficiency. Current approaches can generally be categorized into techniques based on \textit{Model Compression} and those based on \textit{Feature Caching}.
\vspace{-3mm}
\paragraph{Model Compression-based Acceleration.}
Model compression techniques include knowledge distillation~\cite{li2024snapfusion}, network pruning~\cite{structural_pruning_diffusion, zhu2024dipgo}, token reduction~\cite{bolya2023tomesd, kim2024tofu, zhang2024tokenpruningcachingbetter, zhang2025sito, cheng2025catpruningclusterawaretoken, saghatchian2025cached}, and quantization~\cite{10377259, shang2023post, kim2025ditto}. Recently, model grafting~\cite{Chandrasegaran2025ExploringDT} for DiTs has also emerged as a promising direction for exploration.
Although effective, these approaches are constrained by an inherent trade-off: a reduction in model size often incurs substantial degradation in output quality. Therefore, the development of compression strategies for such methods must be painstakingly precise.

\vspace{-3mm}
\paragraph{Feature Caching-based Acceleration.}
Feature Caching has gained prominence due to its training-free nature and strong empirical performance. Initially introduced in U-Net-based diffusion models~\cite{li2023FasterDiffusion, ma2024deepcache}, caching mechanisms have been adapted to address the high computational cost of Diffusion Transformers (DiTs). Recent advances include FORA~\cite{selvaraju2024fora} and $\Delta$-DiT~\cite{chen2024delta-dit}, reusing attention and MLP representations, DiTFastAttn~\cite{yuan2024ditfastattn} and PAB~\cite{zhao2024PAB}, reducing redundancy in self-attention across spatial, temporal, and conditional dimensions. The ToCa series~\cite{zou2024accelerating, zou2024DuCa, Liu2025SpeCa, Zheng2025Compute,TaylorSeer2025} introduces dynamic feature updates to mitigate information loss, with TaylorSeer~\cite{TaylorSeer2025} proposing a `cache-then-forecast' paradigm to significantly reduce cache loss. Additional adaptive strategies include timestep-aware caching in L2C~\cite{ma2024l2c}, TeaCache~\cite{liu2024timestep}, AdaCache~\cite{Kahatapitiya2024AdaptiveCF}, and SpeCa~\cite{Liu2025SpeCa}, as well as methods like FasterCache~\cite{Lv2024FasterCacheTV} with its CFG-Cache strategy. Other notable contributions are EOC's~\cite{qiu2025acceleratingdiffusiontransformererroroptimized} error optimization, UniCP's~\cite{sun2025unicpunifiedcachingpruning} unified caching-pruning framework, RAS's~\cite{liu2025regionadaptivesamplingdiffusiontransformers} region-adaptive sampling, and MagCache's~\cite{Ma2025MagCacheFV} magnitude-aware strategy. The technique has also been extended to tasks such as image editing~\cite{yanEEditRethinkingSpatial2025}.

However, due to its training-free nature, traditional feature caching methods highly rely on tailored, handcrafted prediction strategies to enhance performance, inevitably leading to the loss of both semantic and fine-grained details. Additionally, conventional caching approaches tend to underperform on models that have undergone step distillation, primarily because the inter-step redundancy is significantly reduced after distillation, as mentioned in Figure \ref{fig:Heading}. 

In this work, we propose DisCa as a further improvement to the traditional feature caching and step-distillation methods, specifically MeanFlow, for its impressive performance. By pruning the aggressive compression components of MeanFlow, we significantly enhanced the stability of the distilled models. Through training, the lost details and semantic information faced by the traditional feature caching are then compensated, thereby reaching our goal of further improving computational efficiency on the distilled model.
\vspace{-2mm}
\section{Method}
\label{sec:method}

\vspace{-1mm}
\subsection{Preliminary}
\vspace{-1mm}
\paragraph{Diffusion Models.} 
Diffusion models \citep{ho2020DDPM} generate structured data by iteratively denoising random noise through a series of denoising steps. Setting $t$ as the timestep and $\beta_t$ the noise variance schedule, $p_\theta(x_{t-1} \mid x_t)$, the conditional probability in the denoising process, can be modeled as:
\begin{equation}
\label{eq: new_reverse_process}\vspace{-1mm}
    \mathcal{N} \left( x_{t-1}; \frac{1}{\sqrt{\alpha_t}} \left( x_t - \frac{1 - \alpha_t}{\sqrt{1 - \bar{\alpha}_t}} \epsilon_\theta(x_t, t) \right), \beta_t \mathbf{I} \right),
   \vspace{-1mm}
\end{equation}
where $\alpha_t = 1 - \beta_t$, $\bar{\alpha}_t = \prod_{i=1}^{T} \alpha_i$, and $T$ denotes the total number of timesteps. Notably, $\epsilon_\theta$, a denoising network parameterized by $\theta$, predicts the noise required for denoising from its input $x_t$ and $t$. The total ${T}$ iterative timesteps required for $\epsilon_\theta$ during image generation represent the majority of computational expense in diffusion models. Recent studies have proved that implementing $\epsilon_\theta$ as a decoder-only transformer often enhances generation quality.
\vspace{-3mm}

\begin{figure*}[t]
\centering
\includegraphics[width=\linewidth]{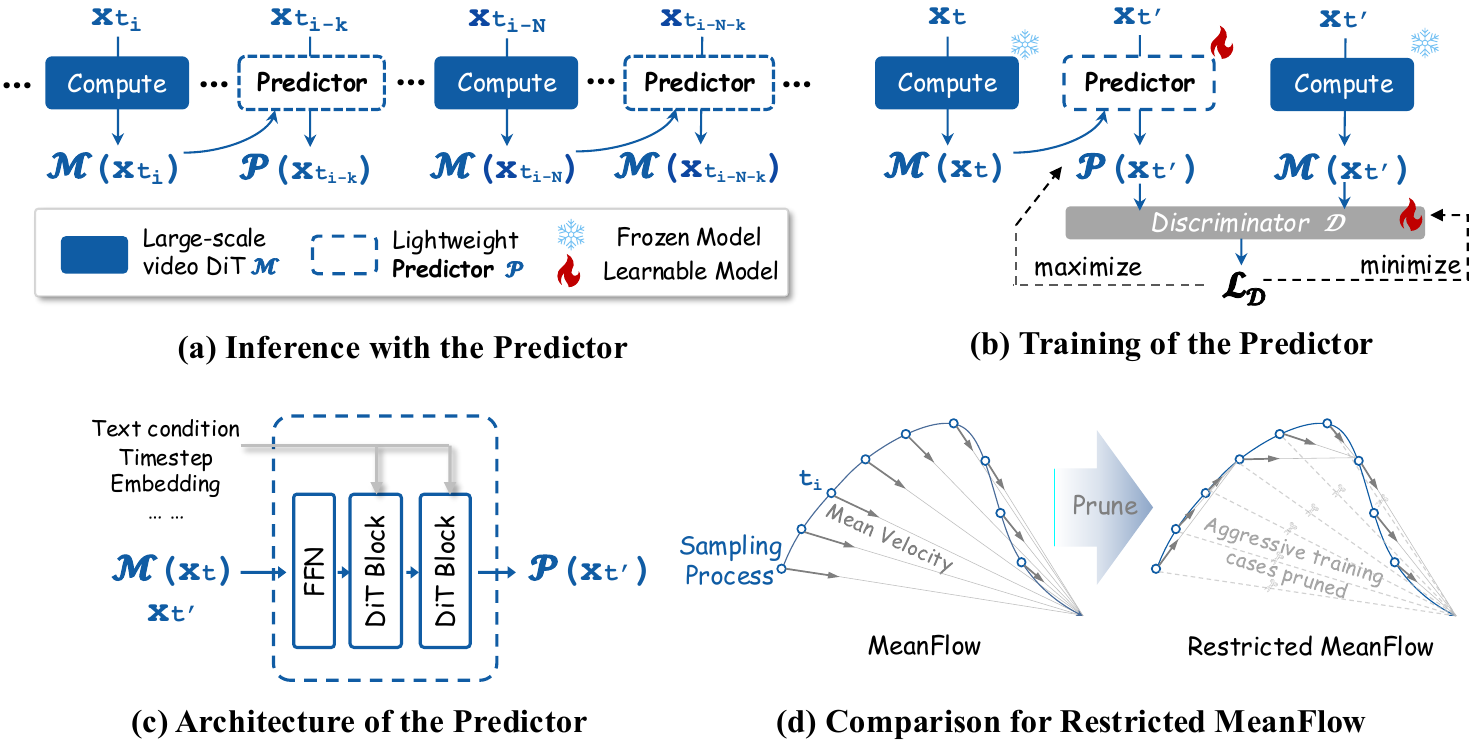}
\vspace{-6mm}
\caption{ \textbf{An overview of Distillation-Compatible Learnable Feature Caching (DisCa).} \textbf{(a) The inference procedure} under the proposed Learnable Feature Caching framework. The lightweight Predictor $\mathcal{P}$ performs multi-step fast inference after a single computation pass through the large-scale DiT $\mathcal{M}$. \textbf{(b) The training process of Predictor.} The cache, initialized by the DiT, is fed into the Predictor as part of the input. The outputs of the Predictor and DiT are passed to the discriminator $\mathcal{D}$, alternating between the objectives of maximizing and minimizing $\mathcal{L}_{\mathcal{D}}$ as part of the adversarial game. \textbf{(c) The lightweight Predictor with two DiT Blocks}, typically constitutes less than 4\% of the total size of the DiT, enabling high-speed and accurate inference. \textbf{(d) The Restricted MeanFlow} is constructed primarily by pruning the components with a high compression ratio in the original MeanFlow, thereby facilitating the learning of the local mean velocity.
}\vspace{-5mm}
\label{fig:Method}
\end{figure*}

\paragraph{Diffusion Transformer Architecture.} 

The Diffusion Transformer (DiT)~\cite{DiT} employs a hierarchical structure, $\mathcal{G} = g_1 \circ g_2 \circ \cdots \circ g_L$, where each block $g_l = \mathcal{F}_{\text{SA}}^l \circ \mathcal{F}_{\text{CA}}^l \circ \mathcal{F}_{\text{MLP}}^l$ comprises self-attention (SA), cross-attention (CA), and multilayer perceptron (MLP) components. The superscript $l \in \{1,2,...,L\}$ denotes the layer index. 
\vspace{-3mm}
\paragraph{Flow Matching.}
Flow Matching \citep{lipman2023FlowMatchingGenerative} is a simple method to train Continuous Normalizing Flows (CNFs), which regresses onto a vector field that generates a target probability density path $p_{t}$. Given two marginal distributions, $q_0(x_0)$ and $q_1(x_1)$ representing data and noise, Flow Matching optimizes a regression objective to learn a CNF for transport between them:
    $\mathbb{E}_{t,p_t(x)}||v_t(x;\theta)-u_t(x)||^{2},$
where $v_t(x;\theta)$ is a parametric vector field for the CNF, and $u_t(x)$ is a target vector field that generates a probability path $p_t$ with the two marginal constraints $p_{t=0}=q_0$ and $p_{t=1}=q_1$. 
\vspace{-3mm}
\paragraph{Na\"ive Feature Caching for Diffusion Models.}
Recent acceleration methods employ \textit{Na\"ive Feature Caching Strategies}~\cite{ma2024deepcache,selvaraju2024fora} in diffusion models by directly reusing computed features across adjacent timesteps.
Specifically, given timesteps $\{t_i,t_{i-1},\dots,t_{i-(N-1)}\}$, features computed at timestep $t_i$ are cached as $\mathcal{C}(x_{t_i}^l) := \{\mathcal{F}(x_{t_i}^l) \}$.
These cached features are then directly reused for subsequent steps: $\mathcal{F}(x_{t_{i-k}}^l):= \mathcal{F}(x_{t_i}^l)$, where $k \in {1, \dots, N-1}$.
While this approach achieves a theoretical $N$-fold speedup by eliminating redundant computations, it suffers from exponential error accumulation as $N$ increases due to neglecting the temporal dynamics of features.
\vspace{-4mm}

\paragraph{TaylorSeer.}
TaylorSeer~\cite{TaylorSeer2025} improves the `cache-then-reuse' framework of na\"ive feature caching to `cache-then-forecast'. By performing a simple forecast, it achieves a significant reduction in cache error, by maintaining a cache at each layer containing the feature's derivatives of multi-orders: $\mathcal{C}(x_t^l):= \{\mathcal{F}(x_t^l), \Delta\mathcal{F}(x_t^l), ..., \Delta^m\mathcal{F}(x_t^l)\}$, and making predictions through a Taylor series expansion:
$\mathcal{F}_{\textrm{pred},m}(x_{t-k}^l) = \mathcal{F}(x_t^l) + \sum_{i=1}^{m} \frac{\Delta^i\mathcal{F}(x_t^l)}{i! \cdot N^i}(-k)^i .$
\vspace{-3mm}
\paragraph{MeanFlow.}
MeanFlow~\cite{Geng2025MeanFF} novelly shifted the training target for few-step DMs from instant velocity to mean velocity in a physically intuitive way. The mean velocity is defined simply:$  (t-r)\vec u(r, t, x_t) = \int_{r}^{t} \vec v(\tau, x_{\tau})\cdot d\tau$, where $\vec u $ is the mean velocity, $\vec v $ is the instant velocity, $t$ and $r$ are the end and the start of the sampled time interval. Although feasible for discrete simulation, integration in the training/distillation process is expensive. After partial derivative transformations in ~\cite{Geng2025MeanFF}, the optimization target is given:
\vspace{-1mm}
\begin{equation}
\label{eq: MeanFlow}
\mathcal{L}(\theta)=\mathbb{E}\left\|u_\theta(x_t,r,t)-\mathrm{sg}(u_\mathrm{tgt})\right\|_2^2,
\vspace{-2mm}
\end{equation}
where $u_{\mathrm{tgt}}=v(x_t,t)-(t-r)\left(v(x_t,t)\partial_x u_\theta+\partial_t u_\theta\right)$, $u_{\mathrm{tgt}}$ is the meanflow target, $\theta$ is the parameter, $u_\theta(x_t,r,t)$ is the predicted mean velocity between $r$ and $t$ generated by MeanFlow model with input noised latent $x_t$. 
\noindent As shown in (\ref{eq: MeanFlow}), the numerical errors introduced by the derivative calculations become more severe for longer sequence generation, making it challenging for large-scale video model training.

\vspace{-1mm}
\subsection{Restricted MeanFlow}
\vspace{-1mm}
The MeanFlow, as mentioned, is designed initially aiming at the target of one-step distillation, so during the distillation, the interval $\mathcal{I}$ between the sampled start time $r$ and end time $t$, $\mathcal{I}=(t-r) \in [0,1]$. Due to the high complexity of large-scale video generation models and the potential numerical errors mentioned, arbitrarily setting the distillation target to `one-step' distillation can be too hard to compress for the aimed large-scale video DiTs. On the contrary, \textit{the highly compressed part of MeanFlow significantly increases the difficulty of training, leading to catastrophic distortions in the generated results}.

Therefore, to achieve high-quality and stable generation results, a conservative sampling strategy is necessary. In cases where the goal is no longer one-step distillation but instead prioritizing quality, the larger-value portions of $\mathcal{I}$, corresponding to the overly aggressive parts in MeanFlow distillation, can be pruned directly. As a local and more stable solution, the Restricted MeanFlow is introduced, building up on MeanFlow with mean velocity interval $\mathcal{I}$ sampled:
\vspace{-3mm}
\begin{equation}
    \label{eq: restricted MeanFlow}
    \mathcal{I}=(t-r) \in [0,\mathcal{R}],
\vspace{-2mm}
\end{equation}
where the restrict factor $\mathcal{R}\in (0,1)$.
\vspace{-1mm}
\subsection{Learnable Feature Caching}
\vspace{-1mm}
Based on the `cache-then-forecast' framework, we introduce a lightweight neural predictor to capture the high-dimensional features in DMs, therefore achieving precise predictions with the strong power of a data-driven method.  
\vspace{-4mm}
\paragraph{Inference Process.}
As mentioned in Figure \ref{fig:Method}, Cached Feature $\mathcal{C}$ is firstly initialized, or refreshed (if this is not the first step), in a full inference step with DM: 
\vspace{-2mm}
\begin{equation}
\label{eq: cache init}
    \mathcal{C}(x_{t_i}) = u(x_{t_i}, r_i, t_i) = \mathcal{M}_{\theta_M}(x_{t_i},r_i,t_i,c_{t_i}), \vspace{-2mm}
\end{equation}
where $\mathcal{M}$ denotes the large-scale DM, with the parameter $\theta_M$, and conditional information vector $c$.

At the coming $N-1$ steps, the lightweight neural Predictor is introduced, rather than full computation:\vspace{-2mm}
\begin{equation}
\label{eq: predict}
    u(x_{t'},t',r') = \mathcal{P}_{\theta_p}(\mathcal{C}, x_{t'}, r', t', c_{t'}), \vspace{-2mm}
\end{equation}
where $\mathcal{P}$ and $\mathcal{C}$ denote the Predictor and cache respectively, with the corresponding paired timesteps in cached steps $(t',r') \in \{(t_{i-1},r_{i-1}),\dots,(t_{i-(N-1),}r_{i-(N-1)})\}$\footnote{For restricted MeanFlow sampling, it is obvious that the average velocity predicted by two adjacent steps should be end-to-end (or contiguous), leading to $r_j = t_{j-1}$ for all sampled $(t,r)$ timestep pairs. }.
\vspace{-4mm}
\paragraph{Predictor Training}
To enable the lightweight Predictor $\mathcal{P}$ to sufficiently learn the feature evolution trend of the large model $\mathcal{M}$ during the sampling process, we innovatively designed a cache-based predictor training scheme.

Firstly, the large-scale DM $\mathcal{M}$ performs one sampling step, obtaining the cache $\mathcal{C}$ as shown in (\ref{eq: cache init}).
 $\mathcal{C}(x_{t}) = \mathcal{M}_{\theta_M}(x_{t},r,t,c_{t}),$
where the $x_t$ is sampled from the noise-data interpolation: $x_t$ = $t x_1+(1-t) x_0$ and $(t,r)$ are the sampled paired timesteps under the restriction (\ref{eq: restricted MeanFlow}).

With the cache $\mathcal{C}$ passed to the learnable lightweight Predictor, the Predictor then  makes its computation as (\ref{eq: predict}):
   $ \mathcal{P}_{\theta_p}(\mathcal{C}, x_{t'}, r', t', c_{t'}),$
where $(t',r')=(t-\Delta, r-\Delta)$, $\Delta$ is a sampled small timestep bias, representing the distance between the full computation step above 
and the current cache step. 
The large-scale DM replicates the predictor sampling process, providing a ground-truth $\mathcal{M}_{\theta_M}(x_{t'},r',t',c_{t'})$. 

In summary, the optimization target can be written as:
\vspace{-2mm}
\begin{equation}
\label{eq: Predictor Training}
\mathcal{L}(\theta_p)=\mathbb{E}\left\|\mathcal{M}_{\theta_M}(x_{t'},r',t')-\mathcal{P}_{\theta_p}(\mathcal{C}, x_{t'}, r', t')\right\|_2^2,
\vspace{-2mm}
\end{equation}
where the cache $\mathcal{C} = \mathcal{M}_{\theta_M}(x_{t},r,t,c_{t})$, and condition vectors $c_t$ and $c_{t'}$ are omitted for simplicity.
\vspace{-3mm}
\paragraph{Generative Adversarial Training.}
To further enhance the prediction performance of the lightweight Predictor and mitigate semantic structure loss and blurriness commonly observed in traditional fature caching, a Generative Adversarial Training scheme, as in ~\cite{Sauer2023AdversarialDD}, is introduced.

Specifically, we employ a Multi-Scale Discriminator based on \textit{Spectral Normalization (SN) and Hinge Loss}. This adversarial setup forces the Predictor to not only minimize error in the pixel space (as with MSE) but also to generate samples rich in high-frequency details and possessing high visual fidelity within the perceptual feature space.

In summary, the final optimization targets with Generative Adversarial loss are as follows:
\vspace{-1mm}
\begin{equation}
\begin{aligned}
\label{eq:d_loss}
\mathcal{L}_{\mathcal{D}}= \mathbb{E} [&\max(0, 1 - \mathcal{D}\circ\mathcal{F}\circ \mathcal{M}_{\theta_M}(x_{t'},r',t')) 
                                  \\&+\max(0, 1 + \mathcal{D}\circ\mathcal{F}\circ \mathcal{P}_{\theta_p}(\mathcal{C}, x_{t'}, r', t')) ],
\end{aligned}
\end{equation} \vspace{-1mm}
\begin{equation}
\begin{aligned}
\label{eq:g_loss}
    \mathcal{L}_{\mathcal{P}} = \mathbb{E}[ &\|\mathcal{M}_{\theta_M}(x_{t'},r',t') - \mathcal{P}_{\theta_p}(\mathcal{C}, x_{t'}, r', t') \|_2^2
    \\+  \lambda &\cdot \max(0, 1 - \mathcal{D}\circ\mathcal{F}\circ \mathcal{P}_{\theta_p}(\mathcal{C}, x_{t'}, r', t'))],
\end{aligned}
\end{equation}
where the $\mathcal{L}_{\mathcal{D}}, \mathcal{L}_{\mathcal{P}}$ are the loss functions for discriminator $\mathcal{D}$ and predictor $\mathcal{P}$, respectively.
Here, $\mathcal{F}$ denotes the feature extractor used before the discriminator $\mathcal{D}$. To enable feature-space adversarial training, the outputs from both the large-scale DM and the lightweight Predictor are passed to $\mathcal{F}$ to obtain their respective multi-level feature representations. Consistent with the methodology in \cite{Sauer2023AdversarialDD}, we utilize the large-scale DM (acting as a pre-trained backbone) as the feature extractor $\mathcal{F}$, and $\lambda$ is the weight for adversarial loss.

\vspace{-3mm}
\paragraph{Predictor Architecture Design.}
The Decoder-Only architecture of DiTs has shown its robust processing capabilities for features, which we aim to continue in the design of the Predictor. Therefore, the Predictor $\mathcal{P}$ is designed by stacking a small number of DiT Blocks, with a size always smaller than $4\%$ of the large-scale model $\mathcal{M}$.

\begin{table*}[htbp]
\centering
\caption{\textbf{Quantitative comparison on Restricted MeanFlow} for HunyuanVideo on VBench.
}
\vspace{-3mm}
\setlength\tabcolsep{5.0pt} 
  \small
  \resizebox{0.98\textwidth}{!}{
  \begin{tabular}{l | c | c  c | c | c c c}
    \toprule
    {\bf Method}  &{\bf CFG} &\multicolumn{3}{c|}{\bf Acceleration} & \multicolumn{3}{c}{\bf VBench Score (\%)} \\
    \cline{3-5}
    \cline{6-8}
    {\bf HunyuanVideo 1.0 [T2V] \citep{sun_hunyuan-large_2024}} & {\bf Distilled}& {\bf Latency(s) $\downarrow$} & {\bf Speed $\uparrow$} & {\bf Full NFE  $\downarrow$}   & \bf Semantic$\uparrow$ & \bf Quality$\uparrow$ & \bf Total$\uparrow$ \\
    \midrule
  
  $\textbf{Original: 50 steps}$ 
                             & \ding{56}&  {1155.3}  & {1.00$\times$} & {$50 \times 2$}   & 73.5\textcolor{gray!70}{\scriptsize (+0.0\%)}& 81.5\textcolor{gray!70}{\scriptsize (+0.0\%)}& 79.9\textcolor{gray!70}{\scriptsize (+0.0\%)}     \\ 

  $\textbf{CFG Distilled: 50 steps}$ 
                           & \ding{52}&  {581.1}  & {1.99$\times$} & {$50$}   & 66.7\textcolor{gray!70}{\scriptsize (-9.3\%)} & 80.6\textcolor{gray!70}{\scriptsize (-1.1\%)} & 77.9\textcolor{gray!70}{\scriptsize (-2.5\%)}     \\ 
\midrule
  $\textbf{MeanFlow: 20 steps}$ 
                           & \ding{52}&  {232.7}  & {4.96$\times$} & {$20$}   & 66.6\textcolor{gray!70}{\scriptsize (+0.0\%)}& 81.8\textcolor{gray!70}{\scriptsize (+0.0\%)} & 78.8\textcolor{gray!70}{\scriptsize (+0.0\%)}     \\ 
  $\textbf{Restricted MeanFlow$(\mathcal{R}=0.4)$}$ 
                          & \ding{52}&  {232.5}  & {4.97$\times$} & {$20$}   & \underline{70.2}\textcolor{gray!70}{\scriptsize (+4.5\%)}& \textbf{82.0}\textcolor{gray!70}{\scriptsize (+0.2\%)}& \textbf{79.7}\textcolor{gray!70}{\scriptsize (+1.1\%)}      \\ 
\rowcolor{gray!20}
  $\textbf{Restricted MeanFlow$(\mathcal{R}=0.2)$}$ 
                          & \ding{52}&  {232.4}  & {4.97$\times$} & {$20$}   & \textbf{70.4}\textcolor[HTML]{0f98b0}{\scriptsize (+5.7\%)}& \underline{81.8}\textcolor[HTML]{0f98b0}{\scriptsize (+0.0\%)}& \underline{79.5}\textcolor[HTML]{0f98b0}{\scriptsize (+0.9\%)}     \\ 
\midrule
  $\textbf{MeanFlow: 10 steps}$ 
                            & \ding{52}&  {119.4}  & {9.68$\times$} & {$10$}   & 60.9\textcolor{gray!70}{\scriptsize (+0.0\%)} & 80.6\textcolor{gray!70}{\scriptsize (+0.0\%)} & 76.7\textcolor{gray!70}{\scriptsize (+0.0\%)}      \\ 

  $\textbf{Restricted MeanFlow$(\mathcal{R}=0.4)$}$ 
                            & \ding{52}&  {119.2}  & {9.69$\times$} & {$10$}   & \underline{67.6}\textcolor{gray!70}{\scriptsize (+11.0\%)} & \underline{81.3} \textcolor{gray!70}{\scriptsize (+0.9\%)}& \underline{78.6} \textcolor{gray!70}{\scriptsize (+2.5\%)}     \\ 
\rowcolor{gray!20}
  $\textbf{Restricted MeanFlow$(\mathcal{R}=0.2)$}$ 
                            & \ding{52}&  {119.3}  & {9.68$\times$} & {$10$}   & \textbf{68.2}\textcolor[HTML]{0f98b0}{\scriptsize (+12.0\%)} & \textbf{81.3} \textcolor[HTML]{0f98b0}{\scriptsize (+0.9\%)}& \textbf{78.7} \textcolor[HTML]{0f98b0}{\scriptsize (+2.9\%)}     \\ 

    \bottomrule
  \end{tabular}}
  
  \label{table:Restriced Meanflow}
\vspace{-6mm}
\end{table*}

\vspace{-3mm}
\paragraph{Memory-Efficient Feature Caching.}
Note that there is no superscript indicating the layer index $l$, which means we no longer maintain the cache composed of one or multiple tensors \textit{for each layer} of the DM; instead, only a single tensor is kept during the inference procedure. 
This primarily benefits from the powerful learning capability of the Predictor, which eliminates the need for a structurally complex cache to provide sufficiently rich information to the training-free solution, as was previously required in TaylorSeer~\cite{TaylorSeer2025}. A Memory-Efficient cache is particularly crucial for the real-world application of high-resolution and long-term video generation models, and simultaneously significantly preserves computational efficiency in distributed parallel environments. We will discuss this further in the following Experiments section and the Appendix.
\vspace{-1mm}
\section{Experiments}
\vspace{-2mm}
\label{sec:experiments}
\subsection{Exeriment Settings}
\vspace{-2mm}
\noindent\textbf{Model Configuration}
The experiments are conducted on HunyuanVideo~\cite{kong2024hunyuanvideo}, a state-of-the-art large-scale video DiT model. Using a checkpoint pre-trained at 540p resolution as the base, distillation and training are carried out on H20. Videos of $704 \times 704$ resolution, 129 frames, and 5 seconds in duration are generated for evaluation. In video generation scenarios involving such high resolution and long time duration, the VRAM pressure becomes crucial. Therefore, a sequence parallel size of 4 is applied. 

\noindent\textbf{Metrics and Evaluation}
For the generated videos, we use VBench~\cite{VBench} for evaluation. VBench comprises 16 sub-dimensions that assess video generation quality from multiple aspects. Among these, 9 aspects constitute the Semantic Score, while the others form the Quality Score. These scores are weighted to produce the Total Score. Given that the Semantic score is more sensitive to distortions and controllability in the generated results, we have selected it as the primary metric, while the others are supplementary references. More information and further experiments on image-to-video tasks are available in the Appendix.

\noindent\textbf{Distillation and Training Configurations}
    Starting from the initial HunyuanVideo with Classifier Free Guidance (CFG), we first perform CFG distillation with a learning rate of $10^{-5}$ to enable the model to directly obtain the result that previously required separate inference with and without CFG in a single forward pass. Subsequently, Restricted MeanFlow distillation is applied sequentially with the same learning rate to complete the step distillation process. Finally, a lightweight predictor is trained with a Generative-Adversarial strategy on the distilled model, with learning rates $10^{-4}, 10^{-2}$ for predictor and discriminator, respectively and a weight of adversarial loss for predictor $\lambda=1.0$ to yield DisCa. More settings are detailed in the Appendix.
\vspace{-3mm}
\subsection{Restricted MeanFlow}
\vspace{-1mm}
\begin{figure}
\centering\includegraphics[width=\linewidth]{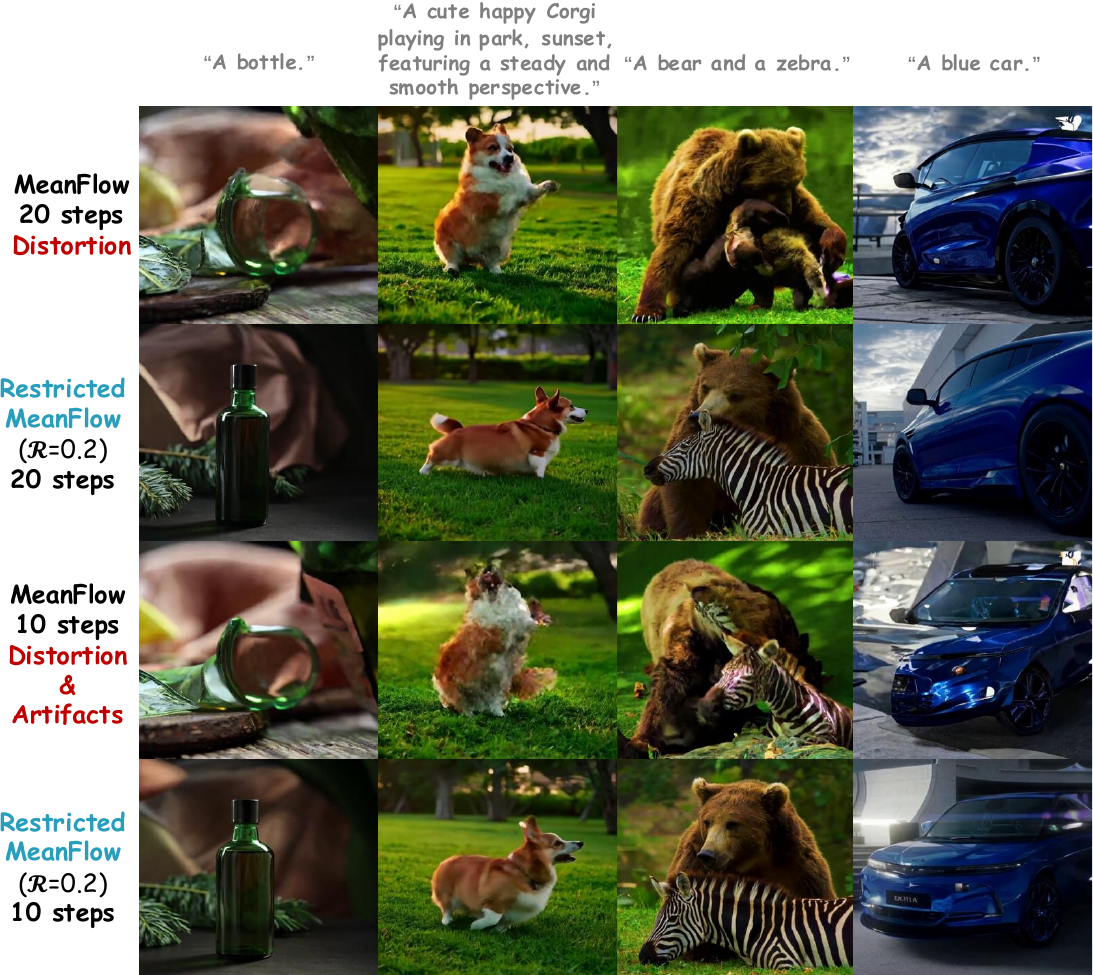}
\vspace{-7mm}
\caption{\textbf{Qualitative Comparison for MeanFlow and the proposed Restrict MeanFlow.} 
In the video generation scenarios for both 20 steps and 10 steps, the MeanFlow method exhibits noticeable distortion and artifacts. In contrast, the Restricted MeanFlow maintains high quality, as indicated by the metrics in Table \ref{table:Restriced Meanflow}.
}
\vspace{-7mm}

\label{fig:Restricted_MFL_vis}
\end{figure}

\noindent\textbf{Quantitative Study}
As shown in Table \ref{table:Restriced Meanflow}, results on both 20 steps and 10 steps demonstrate that the proposed Restricted MeanFlow significantly outperforms the original one-step-aimed MeanFlow~\cite{Geng2025MeanFF}, even surpassing the 50-step model trained with CFG distillation. 
Taking the $\mathcal{R}=0.2$ case as an example, the proposed Restricted MeanFlow exhibits a substantial advantage in the semantic score. For the 20-step generation, it surpasses the original MeanFlow by 5.4\%, and for the more aggressive 10-step generation, this lead increases to a remarkable 12.0\%. It also shows superior performance in the quality score. Specifically, the massive success in the semantic score~\cite{VBench} indicates that the proposed Restricted MeanFlow achieves a significant improvement in terms of controllable generation and mitigating artifacts/distortions compared to the original MeanFlow.

\begin{table*}[htbp]
\centering
\caption{\textbf{Quantitative comparison on different accleration methods} for HunyuanVideo on VBench.
}
\vspace{-3mm}
\setlength\tabcolsep{5.0pt} 
  \small
  \resizebox{0.98\textwidth}{!}{
  \begin{tabular}{l | c | c  c | c | c c c}
    \toprule
    {\bf Method} &{\bf CFG} &\multicolumn{3}{c|}{\bf Efficiency} & \multicolumn{3}{c}{\bf VBench Score (\%)} \\
    \cline{3-5}
    \cline{6-8}
    {\bf HunyuanVideo 1.0 [T2V] \citep{sun_hunyuan-large_2024}} & {\bf Distilled}& {\bf Latency(s) $\downarrow$} & {\bf Speed $\uparrow$} & {\bf Peak VRAM  $\downarrow$}   & \bf Semantic$\uparrow$ & \bf Quality$\uparrow$ & \bf Total$\uparrow$ \\
    \midrule
  
  $\textbf{Original: 50 steps}$ 
                            & \ding{56}&  {1155.3}  & {1.00$\times$} & {99.23GB}   & 73.5\textcolor{gray!70}{\scriptsize (+0.0\%)}& 81.5\textcolor{gray!70}{\scriptsize (+0.0\%)}& 79.9\textcolor{gray!70}{\scriptsize (+0.0\%)}     \\ 

  $\textbf{CFG Distilled: 50 steps}$ 
                            & \ding{52}&  {581.1}  & {1.99$\times$} & {97.21GB}   & 66.7\textcolor{gray!70}{\scriptsize (-9.3\%)} & 80.6\textcolor{gray!70}{\scriptsize (-1.1\%)} & 77.9\textcolor{gray!70}{\scriptsize (-2.5\%)}     \\ 
\midrule
  $\textbf{Original: 10 steps}$ 
                            & \ding{56}&  {234.7}  & {4.92$\times$} & {99.23GB}   & 57.6\textcolor{gray!70}{\scriptsize (-21.6\%)}& 75.8\textcolor{gray!70}{\scriptsize (-7.0\%)}& 72.2\textcolor{gray!70}{\scriptsize (-9.6\%)}     \\ 
  $\textbf{CFG Distilled: 20 steps}$ 
                             & \ding{52}&  {234.4}  & {4.93$\times$} & {97.21GB}   & 64.1\textcolor{gray!70}{\scriptsize (-12.8\%)}& 79.6\textcolor{gray!70}{\scriptsize (-2.3\%)}& 76.5\textcolor{gray!70}{\scriptsize (-4.3\%)}     \\ 
  $\textbf{$\Delta$-DiT$(N=5)$}$ ~\cite{chen2024delta-dit}
                             & \ding{52}&  {306.7}  & {3.77$\times$} & {97.68GB}   &  60.0\textcolor{gray!70}{\scriptsize (-18.4\%)}& 76.7\textcolor{gray!70}{\scriptsize (-5.9\%)}& 73.3\textcolor{gray!70}{\scriptsize (-8.3\%)}     \\  
  $\textbf{PAB$(N=5)$}$ ~\cite{zhao2024PAB}
                             & \ding{52}&  {216.5}  & {5.34$\times$} & {121.3GB}   &  53.4 \textcolor{gray!70}{\scriptsize (-27.3\%)}& 73.1\textcolor{gray!70}{\scriptsize (-10.3\%)}& 69.2\textcolor{gray!70}{\scriptsize (-13.4\%)}     \\  
  $\textbf{TeaCache$(l=0.15)$}$ ~\cite{liu2024timestep}
                            & \ding{52}&  {237.6}  & {5.00$\times$} & {97.70GB}   &  65.5\textcolor{gray!70}{\scriptsize (-10.9\%)}& 80.3\textcolor{gray!70}{\scriptsize (-1.5\%)}& 77.4\textcolor{gray!70}{\scriptsize (-3.1\%)}     \\
  $\textbf{FORA$(N=3)$}$ ~\cite{selvaraju2024fora}
                            & \ding{52}&  {265.7}  & {4.35$\times$} & {124.6GB}   & 63.9\textcolor{gray!70}{\scriptsize (-13.1\%)} &79.7\textcolor{gray!70}{\scriptsize (-2.2\%)}& 76.6\textcolor{gray!70}{\scriptsize (-4.1\%)}     \\    
  $\textbf{TaylorSeer$(N=3,O=1)$}$ ~\cite{TaylorSeer2025}
                             & \ding{52}&  {268.3}  & {4.31$\times$} & {130.7GB}   &  65.2\textcolor{gray!70}{\scriptsize (-11.3\%)}& 80.6\textcolor{gray!70}{\scriptsize (-1.1\%)}& 77.5\textcolor{gray!70}{\scriptsize (-3.0\%)}     \\  
  $\textbf{MeanFlow: 20 steps}$ ~\cite{Geng2025MeanFF}
                             & \ding{52}&  {232.7}  & {4.96$\times$} & {97.21GB}   & 66.6\textcolor{gray!70}{\scriptsize (-9.4\%)}& 81.8\textcolor{gray!70}{\scriptsize (+0.4\%)} & 78.8\textcolor{gray!70}{\scriptsize (-1.4\%)}     \\ 
  $\textbf{Restricted MeanFlow: 20 steps}$[Ours]
                            & \ding{52}&  {232.4}  & {4.97$\times$} & {97.21GB}   & \underline{70.4}\textcolor{gray!70}{\scriptsize (-4.2\%)}& \underline{81.8}\textcolor{gray!70}{\scriptsize (+0.4\%)}& \underline{79.5}\textcolor{gray!70}{\scriptsize (-0.5\%)}     \\ 
\rowcolor{gray!20}
  $\textbf{DisCa$(\mathcal{R}=0.2, N=2)$}$ [Ours]
                            & \ding{52}&  {152.8}  & {7.56$\times$} & {97.64GB}   & \textbf{70.8}\textcolor[HTML]{0f98b0}{\scriptsize (-3.7\%)}& \textbf{81.9}\textcolor[HTML]{0f98b0}{\scriptsize (+0.5\%)}& \textbf{79.7}\textcolor[HTML]{0f98b0}{\scriptsize (-0.3\%)}     \\ 
\midrule
  $\textbf{CFG Distilled: 10 steps}$ 
                            & \ding{52}&  {119.7}  & {9.65$\times$} & {97.21GB}   & 59.0 \textcolor{gray!70}{\scriptsize (-19.7\%)}& 76.8\textcolor{gray!70}{\scriptsize (-4.7\%)}& 73.2 \textcolor{gray!70}{\scriptsize (-8.4\%)}     \\ 
  $\textbf{$\Delta$-DiT$(N=8)$}$ ~\cite{chen2024delta-dit}
                            & \ding{52}&  {253.7}  & {4.55$\times$} & {97.68GB}   &  42.7\textcolor{gray!70}{\scriptsize (-41.9\%)}& 70.9\textcolor{gray!70}{\scriptsize (-13.0\%)}& 65.2\textcolor{gray!70}{\scriptsize (-18.4\%)}     \\  

  $\textbf{PAB$(N=8)$}$ ~\cite{zhao2024PAB}
                            & \ding{52}&  {178.8}  & {6.46$\times$} & {121.3GB}   &  56.3\textcolor{gray!70}{\scriptsize (-23.4\%)}& 76.1\textcolor{gray!70}{\scriptsize (-6.6\%)}& 72.1\textcolor{gray!70}{\scriptsize (-9.8\%)}     \\  
  $\textbf{TeaCache$(l=0.4)$}$ ~\cite{liu2024timestep}
                            & \ding{52}&  {125.3}  & {9.22$\times$} & {97.70GB}   &  62.1\textcolor{gray!70}{\scriptsize (-15.5\%)}& 78.7\textcolor{gray!70}{\scriptsize (-3.4\%)}& 75.4\textcolor{gray!70}{\scriptsize (-5.6\%)}     \\  
  $\textbf{FORA$(N=6)$}$ ~\cite{selvaraju2024fora}
                            & \ding{52}&  {144.2}  & {8.01$\times$} & {124.6GB}   & 57.5\textcolor{gray!70}{\scriptsize (-21.8\%)}& 76.4\textcolor{gray!70}{\scriptsize (-6.3\%)}& 72.6\textcolor{gray!70}{\scriptsize (-9.1\%)}     \\        
  $\textbf{TaylorSeer$(N=6,O=1)$}$ ~\cite{TaylorSeer2025}
                           & \ding{52}&  {166.0}  & {6.96$\times$} & {130.7GB}   &  63.7\textcolor{gray!70}{\scriptsize (-13.3\%)}& 79.9\textcolor{gray!70}{\scriptsize (-2.0\%)}& 76.7\textcolor{gray!70}{\scriptsize (-4.0\%)}     \\  
  $\textbf{Restricted MeanFlow: 9 steps}$[Ours]
                          & \ding{52}&  {108.3}  & {10.7$\times$} & {97.21GB}   & 67.8\textcolor{gray!70}{\scriptsize (-7.8\%)} & {81.0} \textcolor{gray!70}{\scriptsize (-0.6\%)}& 78.4 \textcolor{gray!70}{\scriptsize (-1.9\%)}     \\ 

\rowcolor{gray!20}
  $\textbf{DisCa$(\mathcal{R}=0.2, N=3)$}$ [Ours]
                           & \ding{52}&  {130.7}  & {8.84$\times$} & {97.64GB}   & \textbf{70.3}\textcolor[HTML]{0f98b0}{\scriptsize (-4.4\%)}& \textbf{81.8}\textcolor[HTML]{0f98b0}{\scriptsize (+0.4\%)}& \textbf{79.5}\textcolor[HTML]{0f98b0}{\scriptsize (-0.5\%)}     \\ 
\rowcolor{gray!20}
  $\textbf{DisCa$(\mathcal{R}=0.2, N=4)$}$ [Ours]
                            & \ding{52}&  {97.7}  & {{11.8}$\times$} & {97.64GB}   & \underline{69.3}\textcolor[HTML]{0f98b0}{\scriptsize (-5.7\%)}& {81.1}\textcolor[HTML]{0f98b0}{\scriptsize (-0.5\%)}& \underline{78.8}\textcolor[HTML]{0f98b0}{\scriptsize (-1.4\%)}     \\ 
    \bottomrule
  \end{tabular}}
  
  \label{table:All Metrics}
\vspace{-5mm}
\end{table*}
\noindent\textbf{Qualitative Study}
As shown in Figure \ref{fig:Restricted_MFL_vis}, the proposed Restricted MeanFlow has made significant improvements under accelerated inference at both 20 steps and 10 steps.
Taking the 20-step example, in the ``A bottle." and ``A bear and a zebra." cases, MeanFlow exhibits clear issues of malformation and collapse (or severe distortion), whereas the proposed Restricted MeanFlow still maintains high-quality generation. In the more aggressive 10-step inference, the malformation issues of MeanFlow become even more severe, producing noticeable artifacts and blending issues in the ``... happy Corgi...'' and ``A blue car.'' examples. In contrast, Restricted MeanFlow continues to guarantee quality, further demonstrating that our proposed Restricted MeanFlow, by pruning the training of the aggressive, hard-to-learn portion, effectively enhances video quality.

\vspace{-1mm}
\subsection{Distillation-Compatible Learnable Cache}
\vspace{-1mm}
\noindent\textbf{Quantitative Study}
As shown in Table \ref{table:All Metrics}, the proposed DisCa achieves results that significantly surpass the comparison methods in both the low acceleration ratio and high acceleration ratio regions.
Taking the high-speed region (bottom row) as an example, arbitrarily reducing the total number of sampling steps to 10 steps leads to a 19.7\% semantic loss, with corresponding decreases of 4.7\% and 8.4\% in the Quality and Total scores, respectively. Simple `caching-and-reuse' schemes like $\Delta$-DiT~\cite{chen2024delta-dit}, PAB~\cite{zhao2024PAB}, and FORA~\cite{selvaraju2024fora} completely collapse under high acceleration ratio conditions, all exhibiting a semantic drop exceeding 20\%, along with a total degradation exceeding 9\%. TeaCache~\cite{liu2024timestep}, due to its adaptive caching design, manages to preserve performance to some extent but still suffers from a semantic score decrease of up to 15.5\%. TaylorSeer~\cite{TaylorSeer2025}, leveraging the `cache-then-forecast' framework, performs the best among the various training-free acceleration schemes. However, due to the significant compression in super-high acceleration ratio scenarios, the training-free method can no longer utilize high-dimensional feature information, resulting in a semantic loss of up to 13.3\% and a total score drop of 4\%. 
The proposed DisCa, conversely, not only outperforms Restricted MeanFlow at 4.97$\times$ acceleration while operating at a higher 7.56$\times$ acceleration, but also maintains near-lossless performance even at 8.84$\times$ speed. Even under a super high acceleration ratio of up to 11.8$\times$, it still guarantees high-quality generation, losing only 5.7\% in semantic score and 0.5\% in quality, with an overall total drop of only 1.4\%. This performance even shows a clear improvement compared to the 50-step CFG distilled model, further demonstrating the strong capability of learnable caching with stable distillation.

Note that DisCa is also memory efficient compared to suboptimal caching methods. This is particularly crucial in the discussed real-world scenario of generating high-resolution, long-duration videos. Even with a sequence-parallel size of 4, methods like TaylorSeer and FORA still require over 120 GB of VRAM. Furthermore, ToCa's inability to support the Efficient Attention framework leads to excessively high VRAM consumption for long sequences. This evidence further proves that DisCa is a more suitable acceleration scheme for practical applications.

\begin{figure*}[h!]
\centering
\includegraphics[width=\linewidth]{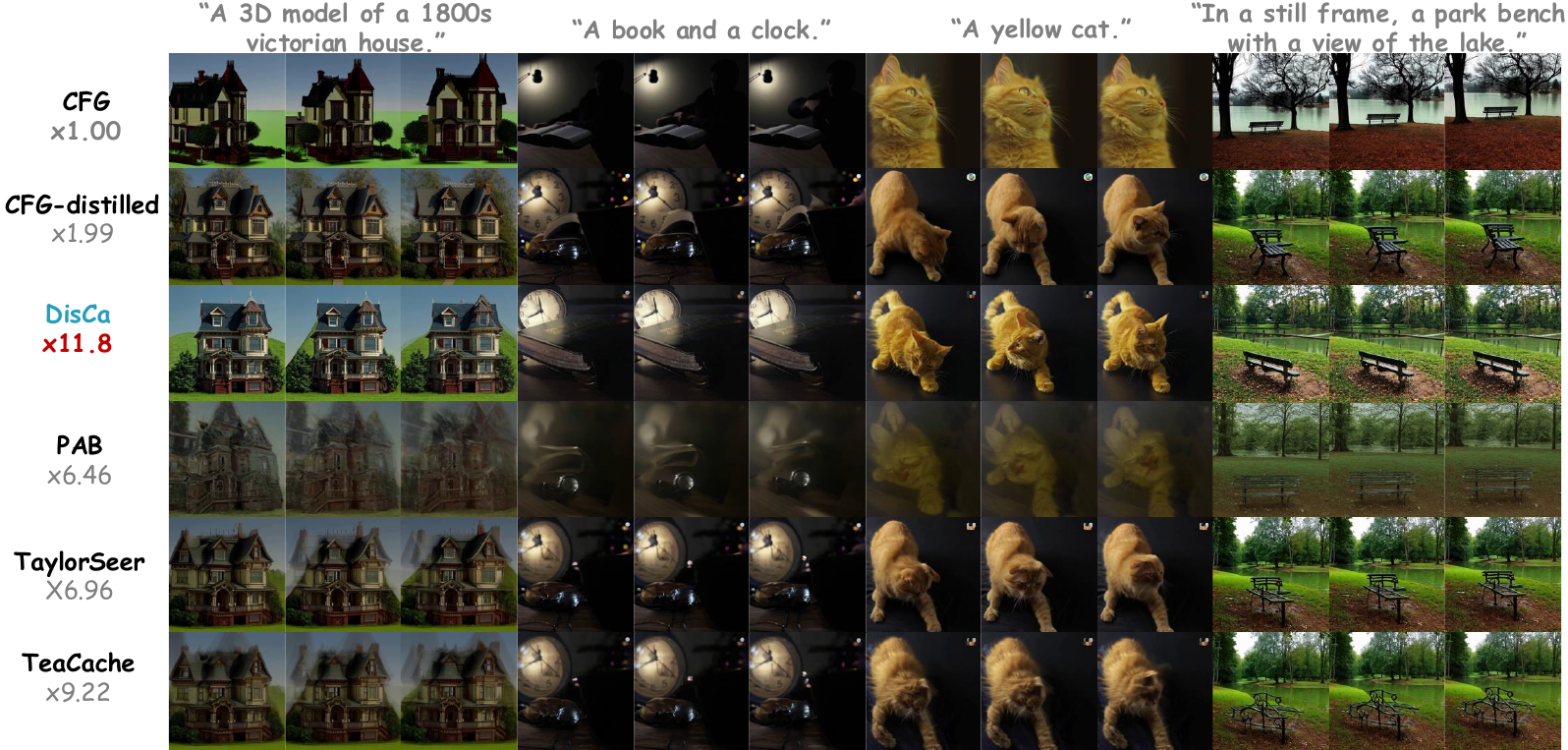}
\vspace{-7mm}
\caption{Visualization of acceleration methods on HunyuanVideo. In the discussed high acceleration ratio scenarios, previous methods exhibit severe degradation, such as malformation and blurring, while \textbf{DisCa} successfully maintains high quality with a $11.8\times$ acceleration.}
\label{fig:Vis-Huanyuan}
\vspace{-6mm}
\end{figure*}

\begin{table}
\centering
\caption{\textbf{Ablation study for Restrict MeanFlow, Learnbale Predictor and GAN Training} in DisCa on HunyuanVideo.
}
\vspace{-3mm}
\setlength\tabcolsep{5.0pt} 

  \small
  \resizebox{0.48\textwidth}{!}{
  \begin{tabular}{c  c  c | c c c}
    \toprule
     {\bf Restricted} &{\bf Learnable} &{\bf GAN} & \multicolumn{3}{c}{\bf VBench Score (\%)} \\
\cline{4-6}
     {\bf MeanFlow} & {\bf Predictor}& {\bf Training }  & \bf Semantic$\uparrow$ & \bf Quality$\uparrow$ & \bf Total$\uparrow$ \\
    \midrule
  
   \ding{52}  & \ding{52}  & \ding{52}   &69.3\textcolor{gray!70}{\scriptsize (+0.0\%)} &81.1\textcolor{gray!70}{\scriptsize (+0.0\%)} &78.7\textcolor{gray!70}{\scriptsize (+0.0\%)} \\ 

   \ding{56}  & \ding{52}  & \ding{52}   &65.2\textcolor{gray!70}{\scriptsize (-5.9\%)} &80.3\textcolor{gray!70}{\scriptsize (-1.0\%)} &77.3\textcolor{gray!70}{\scriptsize (-1.8\%)}  \\ 

   \ding{52}  & \ding{56}  & ---   &67.3\textcolor{gray!70}{\scriptsize (-2.9\%)} &80.5\textcolor{gray!70}{\scriptsize (-0.7\%)} &77.9\textcolor{gray!70}{\scriptsize (-1.0\%)}   \\ 
   
   \ding{52}  & \ding{52}  & \ding{56}   &68.5\textcolor{gray!70}{\scriptsize (-1.2\%)} &81.0\textcolor{gray!70}{\scriptsize (-0.1\%)} &78.5\textcolor{gray!70}{\scriptsize (-0.3\%)}   \\ 
\bottomrule
  \end{tabular}}
  
  \label{table:Ablation Study}
\vspace{-7mm}
\end{table}

\begin{figure}
\centering\includegraphics[width=\linewidth]{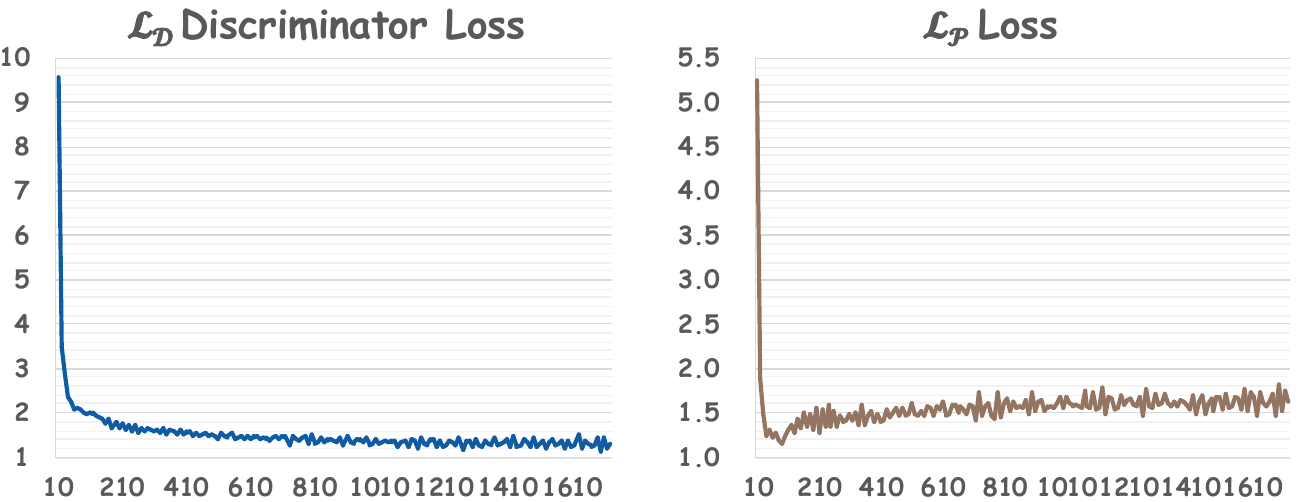}
\vspace{-7mm}
\caption{\textbf{Loss curve during the Generative-Adversarial training process.} The discriminator $\mathcal{D}$ and predictor $\mathcal{P}$ exhibit a stable adversarial dynamic, enhancing the generating capability.
}
\vspace{-6mm}

\label{fig:GAN_Loss}
\end{figure}

\noindent\textbf{Qualitative Study}
Figure \ref{fig:Vis-Huanyuan} demonstrates the inference performance of various acceleration methods on HunyuanVideo. Clearly, the proposed DisCa surpasses previous acceleration methods by an overwhelming margin.

Specifically, PAB~\cite{zhao2024PAB} exhibits noticeable blurring and artifacts across almost all cases. TaylorSeer~\cite{TaylorSeer2025} and TeaCache~\cite{liu2024timestep}, meanwhile, display malformation and distortion in the ``... victorian house." ``... the lake." and ``A book and a clock." examples, with clear detail degradation also evident in the ``A yellow cat." case. In contrast, the proposed DisCa not only achieves an acceleration that far exceeds all comparison methods but also retains rich and diverse details while presenting clear and explicit structural information.

\vspace{-1mm}
\subsection{Ablation Study}
\vspace{-2mm}
We conducted a series of experiments to perform an ablation study on various aspects of the proposed Distillation-Compatible Learnable Feature Caching scheme, further supporting our claims.
As shown in Table \ref{table:Ablation Study}, training the cache scheme on the original MeanFlow rather than the Restricted MeanFlow leads to the most pronounced degradation, resulting in a 5.9\% semantic score drop. This corresponds to a \textit{completely unacceptable malformation} in the generated output, a 1.0\% decrease in the Quality score, and a 1.8\% drop in the total score. This result demonstrates that pruning the overly aggressive portion during distillation is crucial for the video generation task.
When accelerating the distilled model using simple training-free caching, we observed a 2.9\% drop in semantic score and a 0.7\% drop in quality score, \textit{incuring unacceptable losses in both semantic fidelity and quality}. Furthermore, without employing GAN Training, the model's semantic capability also shows a notable decline, with the semantic score decreasing by 1.2\%, further demonstrating the importance of generative adversarial training for the predictor's semantic ability.

Figure \ref{fig:GAN_Loss} displays the training losses $\mathcal{L}_{\mathcal{D}}$ and $\mathcal{L}_{\mathcal{P}}$, for the discriminator and predictor, respectively, during the GAN training process. Both losses first experience a rapid descent during the initial training phase, and subsequently engage in a stable, long-term adversarial dynamic to improve quality.
\vspace{-6mm}
\section{Conclusion}
\label{sec:conclusion}
\vspace{-2mm}
Traditional Feature Caching methods typically rely on direct reuse or simple interpolation, resulting in significant quality degradation under high compression scenarios. Motivated by the fact that the feature evolution of the diffusion model involves rich, high-dimensional information, DisCa firstly proposes overcoming this dilemma by introducing a lightweight, learnable neural predictor to fully utilize the high-dimensional information. We then apply this innovation with our proposed Restricted MeanFlow scheme, which is designed for the stable compression of large-scale video models, successfully increasing the acceleration ratio to $11.8\times$ with high generation quality, offering a novel acceleration solution distinct from conventional wisdom.

\section{Acknowledgment}
We would also like to express our gratitude to Yang Li, Patrol Li, Jianbing Wu, and Xiao He from the Tencent Hunyuan Foundation Model Team for their invaluable support with the underlying codebase. This work was supported by the CCF-Tencent Rhino-Bird Funds.
{
    \small
    \bibliographystyle{ieeenat_fullname}
    \bibliography{main}

@string(CVPR= {IEEE Conf. Comput. Vis. Pattern Recog.})

@string(ICCV= {Int. Conf. Comput. Vis.})

@string(ICLR = {Int. Conf. Learn. Represent.})

@string(AAAI = {AAAI})

@string(CVPR  = {CVPR})

@string(ICCV  = {ICCV})

@string(ICLR  = {ICLR})

@article{structural_pruning_diffusion,
  title        = {Structural Pruning for Diffusion Models},
  author       = {Fang, Gongfan and Ma, Xinyin and Wang, Xinchao},
  year         = 2023,
  journal      = {arXiv preprint arXiv:2305.10924}
}

@article{salimans2022progressive,
  title        = {Progressive distillation for fast sampling of diffusion models},
  author       = {Salimans, Tim and Ho, Jonathan},
  year         = 2022,
  journal      = {arXiv preprint arXiv:2202.00512}
}

@article{selvaraju2024fora,
  title        = {FORA: Fast-Forward Caching in Diffusion Transformer Acceleration},
  author       = {Selvaraju, Pratheba and Ding, Tianyu and Chen, Tianyi and Zharkov, Ilya and Liang, Luming},
  year         = 2024,
  journal      = {arXiv preprint arXiv:2407.01425}
}

@inproceedings{ma2024deepcache,
  title        = {Deepcache: Accelerating diffusion models for free},
  author       = {Ma, Xinyin and Fang, Gongfan and Wang, Xinchao},
  year         = 2024,
  booktitle    = {Proceedings of the IEEE/CVF Conference on Computer Vision and Pattern Recognition},
  pages        = {15762--15772}
}

@article{li2023FasterDiffusion,
  title        = {Faster diffusion: Rethinking the role of unet encoder in diffusion models},
  author       = {Li, Senmao and Hu, Taihang and Khan, Fahad Shahbaz and Li, Linxuan and Yang, Shiqi and Wang, Yaxing and Cheng, Ming-Ming and Yang, Jian},
  year         = 2023,
  journal      = {arXiv preprint arXiv:2312.09608}
}

@article{chen2024delta-dit,
  title        = {$\Delta$-DiT: A Training-Free Acceleration Method Tailored for Diffusion Transformers},
  author       = {Chen, Pengtao and Shen, Mingzhu and Ye, Peng and Cao, Jianjian and Tu, Chongjun and Bouganis, Christos-Savvas and Zhao, Yiren and Chen, Tao},
  year         = 2024,
  journal      = {arXiv preprint arXiv:2406.01125}
}

@article{ma2024l2c,
  title        = {Learning-to-Cache: Accelerating Diffusion Transformer via Layer Caching},
  author       = {Ma, Xinyin and Fang, Gongfan and Mi, Michael Bi and Wang, Xinchao},
  year         = 2024,
  journal      = {arXiv preprint arXiv:2406.01733}
}

@article{zhao2024PAB,
  title        = {Real-Time Video Generation with Pyramid Attention Broadcast},
  author       = {Zhao, Xuanlei and Jin, Xiaolong and Wang, Kai and You, Yang},
  year         = 2024,
  journal      = {arXiv preprint arXiv:2408.12588}
}

@inproceedings{bolya2023tomesd,
  title        = {Token merging for fast stable diffusion},
  author       = {Bolya, Daniel and Hoffman, Judy},
  year         = 2023,
  booktitle    = {Proceedings of the IEEE/CVF conference on computer vision and pattern recognition},
  pages        = {4599--4603}
}

@inproceedings{songDDIM,
  title        = {Denoising Diffusion Implicit Models},
  author       = {Song, Jiaming and Meng, Chenlin and Ermon, Stefano},
  year         = 2021,
  booktitle    = {International Conference on Learning Representations}
}

@inproceedings{peebles2023dit,
  title        = {Scalable diffusion models with transformers},
  author       = {Peebles, William and Xie, Saining},
  year         = 2023,
  booktitle    = {Proceedings of the IEEE/CVF International Conference on Computer Vision},
  pages        = {4195--4205}
}

@inproceedings{chen2023pixartalpha,
  title        = {PixArt-$\alpha$: Fast Training of Diffusion Transformer for Photorealistic Text-to-Image Synthesis},
  author       = {Junsong Chen and Jincheng Yu and Chongjian Ge and Lewei Yao and Enze Xie and Yue Wu and Zhongdao Wang and James Kwok and Ping Luo and Huchuan Lu and Zhenguo Li},
  year         = 2024,
  booktitle    = {International Conference on Learning Representations}
}

@misc{chen2024pixartsigma,
  title        = {PixArt-$\Sigma$: Weak-to-Strong Training of Diffusion Transformer for 4K Text-to-Image Generation},
  author       = {Junsong Chen and Chongjian Ge and Enze Xie and Yue Wu and Lewei Yao and Xiaozhe Ren and Zhongdao Wang and Ping Luo and Huchuan Lu and Zhenguo Li},
  year         = 2024,
  eprint       = {2403.04692},
  archiveprefix = {arXiv},
  primaryclass = {cs.CV}
}

@software{opensora,
  title        = {Open-Sora: Democratizing Efficient Video Production for All},
  author       = {Zangwei Zheng and Xiangyu Peng and Tianji Yang and Chenhui Shen and Shenggui Li and Hongxin Liu and Yukun Zhou and Tianyi Li and Yang You},
  year         = 2024,
  month        = {March},
  url          = {https://github.com/hpcaitech/Open-Sora}
}

@inproceedings{rombach2022SD,
  title        = {High-resolution image synthesis with latent diffusion models},
  author       = {Rombach, Robin and Blattmann, Andreas and Lorenz, Dominik and Esser, Patrick and Ommer, Bj{\"o}rn},
  year         = 2022,
  booktitle    = {Proceedings of the IEEE/CVF conference on computer vision and pattern recognition},
  pages        = {10684--10695}
}

@article{ho2020DDPM,
  title        = {Denoising diffusion probabilistic models},
  author       = {Ho, Jonathan and Jain, Ajay and Abbeel, Pieter},
  year         = 2020,
  journal      = {Advances in neural information processing systems},
  volume       = 33,
  pages        = {6840--6851}
}

@inproceedings{sohl2015deep,
  title        = {Deep unsupervised learning using nonequilibrium thermodynamics},
  author       = {Sohl-Dickstein, Jascha and Weiss, Eric and Maheswaranathan, Niru and Ganguli, Surya},
  year         = 2015,
  booktitle    = {International conference on machine learning},
  pages        = {2256--2265},
  organization = {PMLR}
}

@article{openai2024sora,
  title        = {Video generation models as world simulators},
  author       = {Tim Brooks and Bill Peebles and Connor Holmes and Will DePue and Yufei Guo and Li Jing and David Schnurr and Joe Taylor and Troy Luhman and Eric Luhman and Clarence Ng and Ricky Wang and Aditya Ramesh},
  year         = 2024,
  url          = {https://openai.com/research/video-generation-models-as-world-simulators}
}

@inproceedings{ronneberger2015unet,
  title        = {U-net: Convolutional networks for biomedical image segmentation},
  author       = {Ronneberger, Olaf and Fischer, Philipp and Brox, Thomas},
  year         = 2015,
  booktitle    = {Medical image computing and computer-assisted intervention--MICCAI 2015: 18th international conference, Munich, Germany, October 5-9, 2015, proceedings, part III 18},
  pages        = {234--241},
  organization = {Springer}
}

@article{blattmann2023SVD,
  title        = {Stable video diffusion: Scaling latent video diffusion models to large datasets},
  author       = {Blattmann, Andreas and Dockhorn, Tim and Kulal, Sumith and Mendelevitch, Daniel and Kilian, Maciej and Lorenz, Dominik and Levi, Yam and English, Zion and Voleti, Vikram and Letts, Adam and others},
  year         = 2023,
  journal      = {arXiv preprint arXiv:2311.15127}
}

@inproceedings{zou2024accelerating, 
  title        = {Accelerating Diffusion Transformers with Token-wise Feature Caching},
  author       = {Zou, Chang and Liu, Xuyang and Liu, Ting and Huang, Siteng and Zhang, Linfeng},
  booktitle    = {Proceedings of the 13th International Conference on Learning Representations (ICLR 2025)},
  year         = {2025},
  url          = {https://openreview.net/forum?id=yYZbZGo4ei},  
  note         = {accepted to ICLR 2025},
  publisher    = {ICLR}
}

@inproceedings{TaylorSeer2025,
  title={From Reusing to Forecasting: Accelerating Diffusion Models with TaylorSeers},
  author={Liu, Jiacheng and Zou, Chang and Lyu, Yuanhuiyi and Chen, Junjie and Zhang{, Linfeng},
  booktitle    = International Conference on Computer Vision, ICCV 2025},
  year         = {2025},
  journal={arXiv preprint arXiv:2503.06923},
  year={2025}
}

@inproceedings{Liu2025SpeCa,
  title        = {SpeCa: Accelerating Diffusion Transformers with Speculative Feature Caching},
  author       = {Liu, Jiacheng and Zou, Chang and Lyu, Yuanhuiyi and Li, Kaixin and Wang, Shaobo and Zhang, Linfeng},
  booktitle    = {Proceedings of the 33rd ACM International Conference on Multimedia (MM '25)},
  year         = {2025},
  pages        = {to appear},
  publisher    = {ACM},
  address      = {Dublin, Ireland},
  month        = {October},
  institution  = {Shanghai Jiao Tong University and Shandong University and University of Electronic Science and Technology of China 
                  and The Hong Kong University of Science and Technology (Guangzhou) 
                  and National University of Singapore and Shandong University}
}

@inproceedings{Zheng2025Compute,
  title        = {{Compute only 16 tokens in one timestep: Accelerating Diffusion Transformers with Cluster-Driven Feature Caching}},
  author       = {Zheng, Zhixin and Wang, Xinyu and Zou, Chang and Wang, Shaobo and Zhang, Linfeng},
  booktitle    = {Proceedings of the 33rd ACM International Conference on Multimedia (MM '25)},
  year         = {2025},
  pages        = {to appear},
  publisher    = {ACM},
  address      = {Dublin, Ireland},
  month        = {October},
  institution  = {
    Shanghai Jiao Tong University and 
    University of Electronic Science and Technology of China and 
    Shandong University
  }
}

@software{open_sora_plan,
  title        = {Open-Sora-Plan},
  author       = {PKU-Yuan Lab and Tuzhan AI etc.},
  year         = 2024,
  month        = apr,
  publisher    = {GitHub},
  doi          = {10.5281/zenodo.10948109},
  url          = {https://doi.org/10.5281/zenodo.10948109}
}

@article{lu2022dpm,
  title        = {Dpm-solver: A fast ode solver for diffusion probabilistic model sampling in around 10 steps},
  author       = {Lu, Cheng and Zhou, Yuhao and Bao, Fan and Chen, Jianfei and Li, Chongxuan and Zhu, Jun},
  year         = 2022,
  journal      = {Advances in Neural Information Processing Systems},
  volume       = 35,
  pages        = {5775--5787}
}

@article{lu2022dpm++,
  title        = {Dpm-solver++: Fast solver for guided sampling of diffusion probabilistic models},
  author       = {Lu, Cheng and Zhou, Yuhao and Bao, Fan and Chen, Jianfei and Li, Chongxuan and Zhu, Jun},
  year         = 2022,
  journal      = {arXiv preprint arXiv:2211.01095}
}

@inproceedings{
zheng2023dpmsolvervF,
title={{DPM}-Solver-v3: Improved Diffusion {ODE} Solver with Empirical Model Statistics},
author={Kaiwen Zheng and Cheng Lu and Jianfei Chen and Jun Zhu},
booktitle={Thirty-seventh Conference on Neural Information Processing Systems},
year={2023},
url={https://openreview.net/forum?id=9fWKExmKa0}
}

@inproceedings{kim2024tofu,
  title        = {Token fusion: Bridging the gap between token pruning and token merging},
  author       = {Kim, Minchul and Gao, Shangqian and Hsu, Yen-Chang and Shen, Yilin and Jin, Hongxia},
  year         = 2024,
  booktitle    = {Proceedings of the IEEE/CVF Winter Conference on Applications of Computer Vision},
  pages        = {1383--1392}
}

@article{li2024snapfusion,
  title        = {Snapfusion: Text-to-image diffusion model on mobile devices within two seconds},
  author       = {Li, Yanyu and Wang, Huan and Jin, Qing and Hu, Ju and Chemerys, Pavlo and Fu, Yun and Wang, Yanzhi and Tulyakov, Sergey and Ren, Jian},
  year         = 2024,
  journal      = {Advances in Neural Information Processing Systems},
  volume       = 36
}

@inproceedings{shang2023post,
  title        = {Post-training quantization on diffusion models},
  author       = {Shang, Yuzhang and Yuan, Zhihang and Xie, Bin and Wu, Bingzhe and Yan, Yan},
  year         = 2023,
  booktitle    = {Proceedings of the IEEE/CVF conference on computer vision and pattern recognition},
  pages        = {1972--1981}
}

@inproceedings{yin2024stepdistillation,
  title        = {One-step diffusion with distribution matching distillation},
  author       = {Yin, Tianwei and Gharbi, Micha{\"e}l and Zhang, Richard and Shechtman, Eli and Durand, Fredo and Freeman, William T and Park, Taesung},
  year         = 2024,
  booktitle    = {Proceedings of the IEEE/CVF Conference on Computer Vision and Pattern Recognition},
  pages        = {6613--6623}
}

@inproceedings{song2023consistency,
  title        = {Consistency Models},
  author       = {Song, Yang and Dhariwal, Prafulla and Chen, Mark and Sutskever, Ilya},
  year         = 2023,
  booktitle    = {International Conference on Machine Learning},
  pages        = {32211--32252},
  organization = {PMLR}
}

@inproceedings{refitiedflow,
  title        = {Flow Straight and Fast: Learning to Generate and Transfer Data with Rectified Flow},
  author       = {Liu, Xingchao and Gong, Chengyue and others},
  year         = 2023,
  booktitle    = {The Eleventh International Conference on Learning Representations}
}

@misc{flux2024,
  title        = {FLUX},
  author       = {Black Forest Labs},
  year         = 2024,
  howpublished = {\url{https://github.com/black-forest-labs/flux}}
}

@inproceedings{
yang2025cogvideox,
title={CogVideoX: Text-to-Video Diffusion Models with An Expert Transformer},
author={Zhuoyi Yang and Jiayan Teng and Wendi Zheng and Ming Ding and Shiyu Huang and Jiazheng Xu and Yuanming Yang and Wenyi Hong and Xiaohan Zhang and Guanyu Feng and Da Yin and Xiaotao Gu and Yuxuan.Zhang and Weihan Wang and Yean Cheng and Bin Xu and Yuxiao Dong and Jie Tang},
booktitle={The Thirteenth International Conference on Learning Representations},
year={2025},
url={https://openreview.net/forum?id=LQzN6TRFg9}
}

@inproceedings{
meng2022on,
title={On Distillation of Guided Diffusion Models},
author={Chenlin Meng and Ruiqi Gao and Diederik P Kingma and Stefano Ermon and Jonathan Ho and Tim Salimans},
booktitle={NeurIPS 2022 Workshop on Score-Based Methods},
year={2022},
url={https://openreview.net/forum?id=6QHpSQt6VR-}
}

@INPROCEEDINGS{10377259,
  author={Li, Xiuyu and Liu, Yijiang and Lian, Long and Yang, Huanrui and Dong, Zhen and Kang, Daniel and Zhang, Shanghang and Keutzer, Kurt},
  booktitle={2023 IEEE/CVF International Conference on Computer Vision (ICCV)}, 
  title={Q-Diffusion: Quantizing Diffusion Models}, 
  year={2023},
  volume={},
  number={},
  pages={17489-17499},
  doi={10.1109/ICCV51070.2023.01608}}

@misc{zhang2024tokenpruningcachingbetter,
      title={Token Pruning for Caching Better: 9 Times Acceleration on Stable Diffusion for Free}, 
      author={Evelyn Zhang and Bang Xiao and Jiayi Tang and Qianli Ma and Chang Zou and Xuefei Ning and Xuming Hu and Linfeng Zhang},
      year={2024},
      eprint={2501.00375},
      archivePrefix={arXiv},
      primaryClass={cs.CV},
      url={https://arxiv.org/abs/2501.00375}, 
}

@misc{zou2024DuCa,
      title={Accelerating Diffusion Transformers with Dual Feature Caching}, 
      author={Chang Zou and Evelyn Zhang and Runlin Guo and Haohang Xu and Conghui He and Xuming Hu and Linfeng Zhang},
      year={2024},
      eprint={2412.18911},
      archivePrefix={arXiv},
      primaryClass={cs.LG},
      url={https://arxiv.org/abs/2412.18911}, 
}

@inproceedings{zhang2025sito,
  title={Training-Free and Hardware-Friendly Acceleration for Diffusion Models via Similarity-based Token Pruning},
  author={Zhang, Evelyn and Tang, Jiayi and Ning, Xuefei and Zhang, Linfeng},
  booktitle={Proceedings of the AAAI Conference on Artificial Intelligence},
  year={2025}
}

@misc{liu2025regionadaptivesamplingdiffusiontransformers,
      title={Region-Adaptive Sampling for Diffusion Transformers}, 
      author={Ziming Liu and Yifan Yang and Chengruidong Zhang and Yiqi Zhang and Lili Qiu and Yang You and Yuqing Yang},
      year={2025},
      eprint={2502.10389},
      archivePrefix={arXiv},
      primaryClass={cs.CV},
      url={https://arxiv.org/abs/2502.10389}, 
}

@misc{liu2024timestep,
    title={Timestep Embedding Tells: It's Time to Cache for Video Diffusion Model},
    author={Feng Liu and Shiwei Zhang and Xiaofeng Wang and Yujie Wei and Haonan Qiu and Yuzhong Zhao and Yingya Zhang and Qixiang Ye and Fang Wan},
    year={2024},
    eprint={2411.19108},
    archivePrefix={arXiv},
    primaryClass={cs.CV}
}

@misc{qiu2025acceleratingdiffusiontransformererroroptimized,
      title={Accelerating Diffusion Transformer via Error-Optimized Cache}, 
      author={Junxiang Qiu and Shuo Wang and Jinda Lu and Lin Liu and Houcheng Jiang and Yanbin Hao},
      year={2025},
      eprint={2501.19243},
      archivePrefix={arXiv},
      primaryClass={cs.CV},
      url={https://arxiv.org/abs/2501.19243}, 
}

@inproceedings{yuan2024ditfastattn,
      title={Di{TF}astAttn: Attention Compression for Diffusion Transformer Models},
      author={Zhihang Yuan and Hanling Zhang and Lu Pu and Xuefei Ning and Linfeng Zhang and Tianchen Zhao and Shengen Yan and Guohao Dai and Yu Wang},
      booktitle={The Thirty-eighth Annual Conference on Neural Information Processing Systems},
      year={2024},
      url={https://openreview.net/forum?id=51HQpkQy3t}
}

@misc{cheng2025catpruningclusterawaretoken,
      title={CAT Pruning: Cluster-Aware Token Pruning For Text-to-Image Diffusion Models}, 
      author={Xinle Cheng and Zhuoming Chen and Zhihao Jia},
      year={2025},
      eprint={2502.00433},
      archivePrefix={arXiv},
      primaryClass={cs.CV},
      url={https://arxiv.org/abs/2502.00433}, 
}

@misc{saghatchian2025cached,
    title={Cached Adaptive Token Merging: Dynamic Token Reduction and Redundant Computation Elimination in Diffusion Model},
    author={Omid Saghatchian and Atiyeh Gh. Moghadam and Ahmad Nickabadi},
    year={2025},
    eprint={2501.00946},
    archivePrefix={arXiv},
    primaryClass={cs.CV}
}

@misc{sun2025unicpunifiedcachingpruning,
      title={UniCP: A Unified Caching and Pruning Framework for Efficient Video Generation}, 
      author={Wenzhang Sun and Qirui Hou and Donglin Di and Jiahui Yang and Yongjia Ma and Jianxun Cui},
      year={2025},
      eprint={2502.04393},
      archivePrefix={arXiv},
      primaryClass={cs.CV},
      url={https://arxiv.org/abs/2502.04393}, 
}

@inproceedings{kim2025ditto,
  author = {Sungbin Kim and Hyunwuk Lee and Wonho Cho and Mincheol Park and Won Woo Ro},
  title = {Ditto: Accelerating Diffusion Model via Temporal Value Similarity},
  booktitle = {Proceedings of the 2025 IEEE International Symposium on High-Performance Computer Architecture (HPCA)},
  year = {2025},
  publisher = {IEEE},
}

@misc{zhu2024dipgo,
    title={DiP-GO: A Diffusion Pruner via Few-step Gradient Optimization},
    author={Haowei Zhu and Dehua Tang and Ji Liu and Mingjie Lu and Jintu Zheng and Jinzhang Peng and Dong Li and Yu Wang and Fan Jiang and Lu Tian and Spandan Tiwari and Ashish Sirasao and Jun-Hai Yong and Bin Wang and Emad Barsoum},
    year={2024},
    eprint={2410.16942},
    archivePrefix={arXiv},
    primaryClass={cs.CV}
}

@misc{VBench,
	title = {{VBench}: {Comprehensive} {Benchmark} {Suite} for {Video} {Generative} {Models}},
	shorttitle = {{VBench}},
	url = {http://arxiv.org/abs/2311.17982},
	doi = {10.48550/arXiv.2311.17982},
	urldate = {2025-02-27},
	publisher = {arXiv},
	author = {Huang, Ziqi and He, Yinan and Yu, Jiashuo and Zhang, Fan and Si, Chenyang and Jiang, Yuming and Zhang, Yuanhan and Wu, Tianxing and Jin, Qingyang and Chanpaisit, Nattapol and Wang, Yaohui and Chen, Xinyuan and Wang, Limin and Lin, Dahua and Qiao, Yu and Liu, Ziwei},
	month = nov,
	year = {2023},
	note = {arXiv:2311.17982 [cs]},
}

@misc{DM,
	title = {Denoising {Diffusion} {Probabilistic} {Models}},
	url = {http://arxiv.org/abs/2006.11239},
	doi = {10.48550/arXiv.2006.11239},
	urldate = {2025-02-27},
	publisher = {arXiv},
	author = {Ho, Jonathan and Jain, Ajay and Abbeel, Pieter},
	month = dec,
	year = {2020},
	note = {arXiv:2006.11239 [cs]},
}

@misc{DiT,
	title = {Scalable {Diffusion} {Models} with {Transformers}},
	url = {http://arxiv.org/abs/2212.09748},
	doi = {10.48550/arXiv.2212.09748},
	urldate = {2025-02-27},
	publisher = {arXiv},
	author = {Peebles, William and Xie, Saining},
	month = mar,
	year = {2023},
	note = {arXiv:2212.09748 [cs]},

}

@misc{li_hunyuan-dit_2024,
	title = {Hunyuan-{DiT}: A Powerful Multi-Resolution Diffusion Transformer with Fine-Grained Chinese Understanding},
	url = {http://arxiv.org/abs/2405.08748},
	doi = {10.48550/arXiv.2405.08748},
	shorttitle = {Hunyuan-{DiT}},

	number = {{arXiv}:2405.08748},
	publisher = {{arXiv}},
	author = {Li, Zhimin and Zhang, Jianwei and Lin,and others},
	urldate = {2025-03-01},
	date = {2024-05-14},
	eprinttype = {arxiv},
	eprint = {2405.08748 [cs]},
}

@misc{sun_hunyuan-large_2024,
	title = {Hunyuan-Large: An Open-Source {MoE} Model with 52 Billion Activated Parameters by Tencent},
	url = {http://arxiv.org/abs/2411.02265},
	doi = {10.48550/arXiv.2411.02265},
	shorttitle = {Hunyuan-Large},

	number = {{arXiv}:2411.02265},
	publisher = {{arXiv}},
	author = {Sun, Xingwu and Chen, Yanfeng and Huang and others},
	urldate = {2025-03-01},
	date = {2024-11-06},
	eprinttype = {arxiv},
	eprint = {2411.02265 [cs]},

}

@article{kong2024hunyuanvideo,
  title={Hunyuanvideo: A systematic framework for large video generative models},
  author={Kong, Weijie and Tian, Qi and Zhang, Zijian and Min, Rox and Dai, Zuozhuo and Zhou, Jin and Xiong, Jiangfeng and Li, Xin and Wu, Bo and Zhang, Jianwei and others},
  journal={arXiv preprint arXiv:2412.03603},
  year={2024}
}

@misc{hunyuanvideo2025,
      title={HunyuanVideo 1.5 Technical Report}, 
      author={Tencent Hunyuan Foundation Model Team},
      year={2025},
      eprint={2511.18870},
      archivePrefix={arXiv},
      primaryClass={cs.CV},
      url={https://arxiv.org/abs/2511.18870}, 
}

@misc{HunyuanImage-2.1,
  title={HunyuanImage 2.1: An Efficient Diffusion Model for High-Resolution (2K) Text-to-Image Generation},
  author={Tencent Hunyuan Team},
  year={2025},
  howpublished={\url{https://github.com/Tencent-Hunyuan/HunyuanImage-2.1}},
}

@misc{wan_wan_2025,
	title = {Wan: Open and Advanced Large-Scale Video Generative Models},
	url = {http://arxiv.org/abs/2503.20314},
	doi = {10.48550/arXiv.2503.20314},
	shorttitle = {Wan},
	
	number = {{arXiv}:2503.20314},
	publisher = {{arXiv}},
	author = {Wan, Team and Wang, Ang and Ai, Baole and Wen, Bin and Mao, Chaojie and Xie, Chen-Wei and Chen, Di and Yu, Feiwu and Zhao, Haiming and Yang, Jianxiao and Zeng, Jianyuan and Wang, Jiayu and Zhang, Jingfeng and Zhou, Jingren and Wang, Jinkai and Chen, Jixuan and Zhu, Kai and Zhao, Kang and Yan, Keyu and Huang, Lianghua and Feng, Mengyang and Zhang, Ningyi and Li, Pandeng and Wu, Pingyu and Chu, Ruihang and Feng, Ruili and Zhang, Shiwei and Sun, Siyang and Fang, Tao and Wang, Tianxing and Gui, Tianyi and Weng, Tingyu and Shen, Tong and Lin, Wei and Wang, Wei and Wang, Wei and Zhou, Wenmeng and Wang, Wente and Shen, Wenting and Yu, Wenyuan and Shi, Xianzhong and Huang, Xiaoming and Xu, Xin and Kou, Yan and Lv, Yangyu and Li, Yifei and Liu, Yijing and Wang, Yiming and Zhang, Yingya and Huang, Yitong and Li, Yong and Wu, You and Liu, Yu and Pan, Yulin and Zheng, Yun and Hong, Yuntao and Shi, Yupeng and Feng, Yutong and Jiang, Zeyinzi and Han, Zhen and Wu, Zhi-Fan and Liu, Ziyu},
	urldate = {2025-07-24},
	date = {2025-04-19},
	eprinttype = {arxiv},
	eprint = {2503.20314 [cs]},
	
}

@misc{yanEEditRethinkingSpatial2025,
  title = {{{EEdit}} : {{Rethinking}} the {{Spatial}} and {{Temporal Redundancy}} for {{Efficient Image Editing}}},
  shorttitle = {{{EEdit}}},
  author = {Yan, Zexuan and Ma, Yue and Zou, Chang and Chen, Wenteng and Chen, Qifeng and Zhang, Linfeng},
  year = {2025},
  number = {arXiv:2503.10270},
  eprint = {2503.10270},
  primaryclass = {cs},
  publisher = {arXiv},
  doi = {10.48550/arXiv.2503.10270},
  archiveprefix = {arXiv}
}

@misc{kahatapitiyaAdaptiveCachingFaster2024,
  title = {Adaptive {{Caching}} for {{Faster Video Generation}} with {{Diffusion Transformers}}},
  author = {Kahatapitiya, Kumara and Liu, Haozhe and He, Sen and Liu, Ding and Jia, Menglin and Zhang, Chenyang and Ryoo, Michael S. and Xie, Tian},
  year = {2024},
  number = {arXiv:2411.02397},
  eprint = {2411.02397},
  primaryclass = {cs},
  publisher = {arXiv},
  doi = {10.48550/arXiv.2411.02397},
  archiveprefix = {arXiv}
}

@article{Lipman2022FlowMF,
  title={Flow Matching for Generative Modeling},
  author={Yaron Lipman and Ricky T. Q. Chen and Heli Ben-Hamu and Maximilian Nickel and Matt Le},
  journal={ArXiv},
  year={2022},
  volume={abs/2210.02747},
  url={https://api.semanticscholar.org/CorpusID:252734897}
}

@article{Yin2023OneStepDW,
  title={One-Step Diffusion with Distribution Matching Distillation},
  author={Tianwei Yin and Michael Gharbi and Richard Zhang and Eli Shechtman and Fr{\'e}do Durand and William T. Freeman and Taesung Park},
  journal={2024 IEEE/CVF Conference on Computer Vision and Pattern Recognition (CVPR)},
  year={2023},
  pages={6613-6623},
  url={https://api.semanticscholar.org/CorpusID:265506768}
}

@article{Salimans2024MultistepDO,
  title={Multistep Distillation of Diffusion Models via Moment Matching},
  author={Tim Salimans and Thomas Mensink and Jonathan Heek and Emiel Hoogeboom},
  journal={ArXiv},
  year={2024},
  volume={abs/2406.04103},
  url={https://api.semanticscholar.org/CorpusID:270285800}
}

@inproceedings{Lu2025SimplifyingSA,
  title={Simplifying, Stabilizing and Scaling Continuous-time Consistency Models},
  author={Cheng Lu and Yang Song},
  booktitle={International Conference on Learning Representations},
  year={2025},
  url={https://api.semanticscholar.org/CorpusID:278497880}
}

@article{Frans2024OneSD,
  title={One Step Diffusion via Shortcut Models},
  author={Kevin Frans and Danijar Hafner and Sergey Levine and Pieter Abbeel},
  journal={ArXiv},
  year={2024},
  volume={abs/2410.12557},
  url={https://api.semanticscholar.org/CorpusID:273375140}
}

@article{Geng2025MeanFF,
  title={Mean Flows for One-step Generative Modeling},
  author={Zhengyang Geng and Mingyang Deng and Xingjian Bai and J. Zico Kolter and Kaiming He},
  journal={ArXiv},
  year={2025},
  volume={abs/2505.13447},
  url={https://api.semanticscholar.org/CorpusID:278769814}
}

@article{Chandrasegaran2025ExploringDT,
  title={Exploring Diffusion Transformer Designs via Grafting},
  author={Keshigeyan Chandrasegaran and Michael Poli and Daniel Y. Fu and Dongjun Kim and Lea M. Hadzic and Manling Li and Agrim Gupta and Stefano Massaroli and Azalia Mirhoseini and Juan Carlos Niebles and Stefano Ermon and Fei-Fei Li},
  journal={ArXiv},
  year={2025},
  volume={abs/2506.05340},
  url={https://api.semanticscholar.org/CorpusID:279243916}
}

@inproceedings{Lv2024FasterCacheTV,
  title={FasterCache: Training-Free Video Diffusion Model Acceleration with High Quality},
  author={Zhengyao Lv and Chenyang Si and Junhao Song and Zhenyu Yang and Yu Qiao and Ziwei Liu and Kwan-Yee K. Wong},
  booktitle    = {Proceedings of the 13th International Conference on Learning Representations (ICLR 2025)},
  year         = {2025},
  year={2024},
  volume={abs/2410.19355},
  url={https://api.semanticscholar.org/CorpusID:273638044}
}

@article{Kahatapitiya2024AdaptiveCF,
  title={Adaptive Caching for Faster Video Generation with Diffusion Transformers},
  author={Kumara Kahatapitiya and Haozhe Liu and Sen He and Ding Liu and Menglin Jia and Michael S. Ryoo and Tian Xie},
  journal={ArXiv},
  year={2024},
  volume={abs/2411.02397},
  url={https://api.semanticscholar.org/CorpusID:273821120}
}

@article{Ma2025MagCacheFV,
  title={MagCache: Fast Video Generation with Magnitude-Aware Cache},
  author={Zehong Ma and Longhui Wei and Feng Wang and Shiliang Zhang and Qi Tian},
  journal={ArXiv},
  year={2025},
  volume={abs/2506.09045},
  url={https://api.semanticscholar.org/CorpusID:279260602}
}

@article{Luo2023LatentCM,
  title={Latent Consistency Models: Synthesizing High-Resolution Images with Few-Step Inference},
  author={Simian Luo and Yiqin Tan and Longbo Huang and Jian Li and Hang Zhao},
  journal={ArXiv},
  year={2023},
  volume={abs/2310.04378},
  url={https://api.semanticscholar.org/CorpusID:263831037}
}

@misc{wu2025qwenimagetechnicalreport,
      title={Qwen-Image Technical Report}, 
      author={Chenfei Wu and Jiahao Li and Jingren Zhou and Junyang Lin and Kaiyuan Gao and Kun Yan and Sheng-ming Yin and Shuai Bai and Xiao Xu and Yilei Chen and Yuxiang Chen and Zecheng Tang and Zekai Zhang and Zhengyi Wang and An Yang and Bowen Yu and Chen Cheng and Dayiheng Liu and Deqing Li and Hang Zhang and Hao Meng and Hu Wei and Jingyuan Ni and Kai Chen and Kuan Cao and Liang Peng and Lin Qu and Minggang Wu and Peng Wang and Shuting Yu and Tingkun Wen and Wensen Feng and Xiaoxiao Xu and Yi Wang and Yichang Zhang and Yongqiang Zhu and Yujia Wu and Yuxuan Cai and Zenan Liu},
      year={2025},
      eprint={2508.02324},
      archivePrefix={arXiv},
      primaryClass={cs.CV},
      url={https://arxiv.org/abs/2508.02324}, 
}

@article{Eskimez2024E2TE,
  title={E2 TTS: Embarrassingly Easy Fully Non-Autoregressive Zero-Shot TTS},
  author={Sefik Emre Eskimez and Xiaofei Wang and Manthan Thakker and Canrun Li and Chung-Hsien Tsai and Zhen Xiao and Hemin Yang and Zirun Zhu and Min Tang and Xu Tan and Yanqing Liu and Sheng Zhao and Naoyuki Kanda},
  journal={2024 IEEE Spoken Language Technology Workshop (SLT)},
  year={2024},
  pages={682-689},
  url={https://api.semanticscholar.org/CorpusID:270738197}
}

@article{chen-etal-2024-f5tts,
      title={F5-TTS: A Fairytaler that Fakes Fluent and Faithful Speech with Flow Matching}, 
      author={Yushen Chen and Zhikang Niu and Ziyang Ma and Keqi Deng and Chunhui Wang and Jian Zhao and Kai Yu and Xie Chen},
      journal={arXiv preprint arXiv:2410.06885},
      year={2024},
}

@article{Nie2025LargeLD,
  title={Large Language Diffusion Models},
  author={Shen Nie and Fengqi Zhu and Zebin You and Xiaolu Zhang and Jingyang Ou and Jun Hu and Jun Zhou and Yankai Lin and Ji-Rong Wen and Chongxuan Li},
  journal={ArXiv},
  year={2025},
  volume={abs/2502.09992},
  url={https://api.semanticscholar.org/CorpusID:276395038}
}

@article{Zhu2025LLaDA1V,
  title={LLaDA 1.5: Variance-Reduced Preference Optimization for Large Language Diffusion Models},
  author={Fengqi Zhu and Rongzheng Wang and Shen Nie and Xiaolu Zhang and Chunwei Wu and Jun Hu and Jun Zhou and Jianfei Chen and Yankai Lin and Ji-Rong Wen and Chongxuan Li},
  journal={ArXiv},
  year={2025},
  volume={abs/2505.19223},
  url={https://api.semanticscholar.org/CorpusID:278905103}
}

@misc{lipman2023FlowMatchingGenerative,
  title = {Flow {{Matching}} for {{Generative Modeling}}},
  author = {Lipman, Yaron and Chen, Ricky T. Q. and {Ben-Hamu}, Heli and Nickel, Maximilian and Le, Matt},
  year = 2023,
  eprint = {2210.02747},
  primaryclass = {cs},
  publisher = {arXiv},
  doi = {10.48550/arXiv.2210.02747},
}

@inproceedings{Sauer2023AdversarialDD,
  title={Adversarial Diffusion Distillation},
  author={Axel Sauer and Dominik Lorenz and A. Blattmann and Robin Rombach},
  booktitle={European Conference on Computer Vision},
  year={2023},
  url={https://api.semanticscholar.org/CorpusID:265466173}
}

@article{Geng2024OmniCacheAU,
  title={OmniCache: A Unified Cache for Efficient Query Handling in LSM-tree Based Key-Value Stores},
  author={Yiyang Geng and Huai Xu and Yanyong Zhang and Fuxin Zhang},
  journal={2024 IEEE International Conference on High Performance Computing and Communications (HPCC)},
  year={2024},
  pages={353-360},
  url={https://api.semanticscholar.org/CorpusID:280043133}
}

@article{Yin2024FromSB,
  title={From Slow Bidirectional to Fast Autoregressive Video Diffusion Models},
  author={Tianwei Yin and Qiang Zhang and Richard Zhang and William T. Freeman and Fr{\'e}do Durand and Eli Shechtman and Xun Huang},
  journal={2025 IEEE/CVF Conference on Computer Vision and Pattern Recognition (CVPR)},
  year={2024},
  pages={22963-22974},
  url={https://api.semanticscholar.org/CorpusID:274610175}
}

@article{Lv2025DCMDC,
  title={DCM: Dual-Expert Consistency Model for Efficient and High-Quality Video Generation},
  author={Zhengyao Lv and Chenyang Si and Tianlin Pan and Zhaoxi Chen and Kwan-Yee K. Wong and Yu Qiao and Ziwei Liu},
  journal={ArXiv},
  year={2025},
  volume={abs/2506.03123},
  url={https://api.semanticscholar.org/CorpusID:279119323}
}

@article{Zhang2025AccVideoAV,
  title={AccVideo: Accelerating Video Diffusion Model with Synthetic Dataset},
  author={Haiyu Zhang and Xinyuan Chen and Yaohui Wang and Xihui Liu and Yunhong Wang and Yu Qiao},
  journal={ArXiv},
  year={2025},
  volume={abs/2503.19462},
  url={https://api.semanticscholar.org/CorpusID:277313782}
}
}

\clearpage
\setcounter{page}{1}
\maketitlesupplementary

\section{Suppelmentary Experiments}

\subsection{Quantitative comparison for Image-to-Video}
As shown in Table \ref{table:i2v}, besides the discussed experiments on \textbf{HunyuanVideo 1.0} (\textbf{text-to-video}), we conducted additional evaluations on the \textbf{image-to-video} task using the current state-of-the-art video model, \textbf{HunyuanVideo 1.5} to verify generalization, evaluating with PSNR, SSIM, and LPIPS. 
We subjected our proposed methods to \textbf{more aggressive settings} for fair comparisons with alternative distillation methods to demonstrate the necessity of our proposed approach. Low-level metrics were calculated on 300 selected high-quality $1\text{K}$ image-text pairs by comparing the accelerated generated images against their corresponding non-accelerated counterparts. For brevity, our proposed Restricted MeanFlow is abbreviated as R-MeanFlow.\vspace{-2mm}
\begin{table}[htbp]
\caption{\textbf{Quantitative comparison on different accleration methods} for HunyuanVideo 1.5 on image-to-video (I2V) task.
\label{table:i2v}}
\vspace{-10pt}
\setlength\tabcolsep{5.0pt} 
\resizebox{.48\textwidth}{!}{
\begin{tabular}{l|cc|ccc}
\toprule
Method     & NFE$\downarrow$      & Latency(s)$\downarrow$   & LPIPS$\downarrow$   & PSNR$\uparrow$      & SSIM$\uparrow$     \\ 
\midrule
\textbf{HunyuanVideo 1.5 [I2V]}      & 50& 778.1 & -    & -    & -    \\
\midrule
TaylorSeer                           & 8 & 123.1 & 0.2050 & 19.34 & 0.7137 \\
MagCache                             & 8 & 127.6 & 0.2093 & 19.03 & 0.7056  \\

AdaCache                             & 8 & 129.4 & 0.2084 & 19.08 & 0.7070 \\

DCM                                  & 8 & 124.5  & 0.2496 & 18.04 & 0.6616\\

\cellcolor{green!8}\textbf{R-MeanFlow($R=0.2$)}    & \cellcolor{green!8}\textbf{8} & \cellcolor{green!8}\textbf{124.5} & \cellcolor{green!8}\textbf{0.1778} & \cellcolor{green!8}\textbf{19.85} & \cellcolor{green!8}\textbf{0.7186} \\
\midrule
TaylorSeer                           & 4 & 79.6 & 0.2470 & 18.74 & 0.6996 \\
MagCache                             & 4 & 64.5 & 0.2416 & 18.73 & 0.7009 \\
AdaCache                             & 4 & 64.4 & 0.2551 & 18.40 & 0.6889 \\
DCM                                  & 4 & \bf 62.3 & 0.2450 & 17.84 & 0.6686\\

\cellcolor{green!8}\textbf{DisCa ($N=2, R=0.2$)}  & \cellcolor{green!8}\textbf{4}& \cellcolor{green!8}{64.2}& \cellcolor{green!8}\textbf{0.1942}& \cellcolor{green!8}\textbf{19.29}& \cellcolor{green!8}\textbf{0.7071}\\
\bottomrule
\end{tabular}}
\vspace{-10pt}
\end{table}

\subsection{User Studies}
\begin{figure}[h!]
\centering
\vspace{-1mm}
\includegraphics[width=\linewidth]{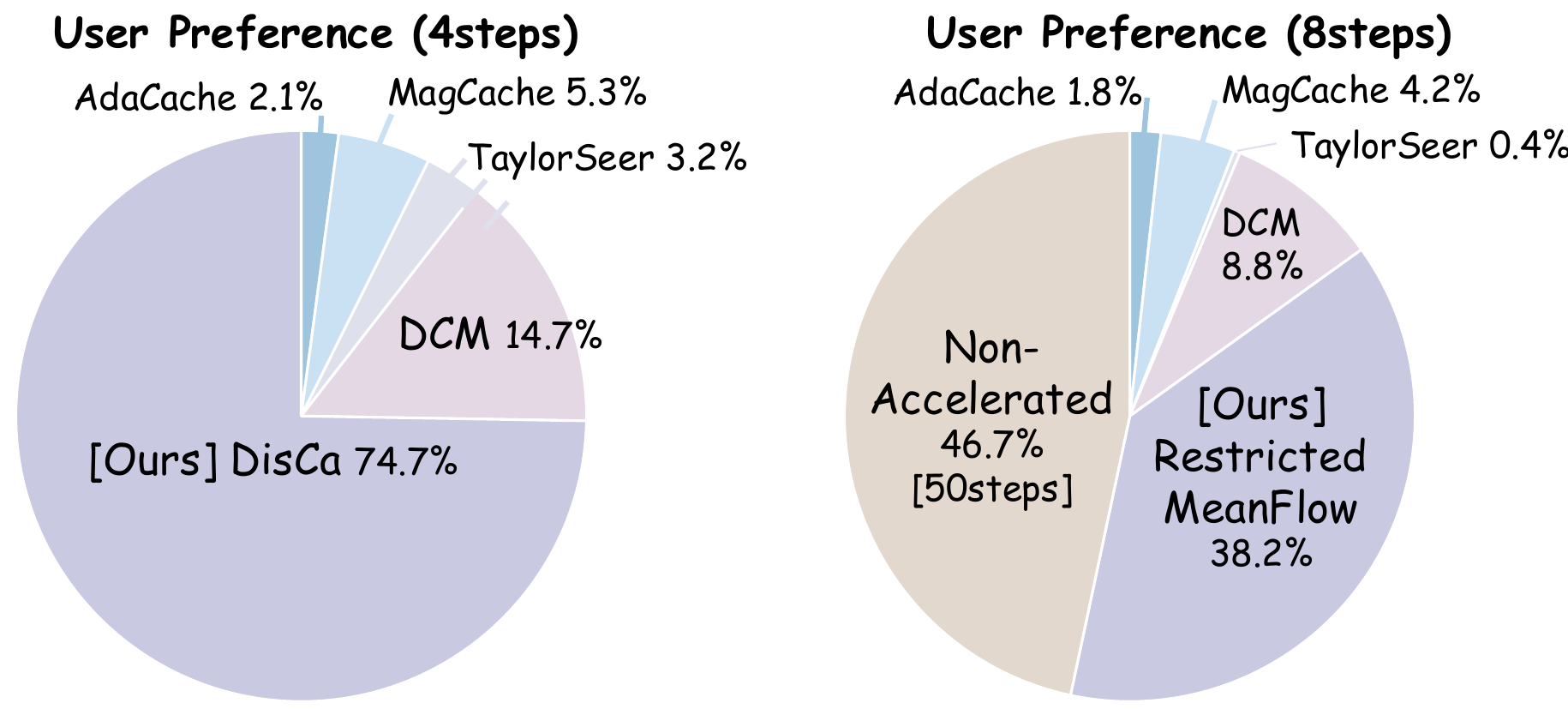}
\vspace{-7mm}
\caption{\textbf{User Studies} for \textbf{DisCa} on HunyuanVideo 1.5 [I2V].\label{fig:User_Study}}
\end{figure}

As shown in Figure \ref{fig:User_Study}, we conducted user studies for 4-step and 8-step inference to provide a more comprehensive comparison, demonstrating our method's clear superiority over the baselines. While a performance gap remains between 8-step inference and the original non-accelerated 50-step generation, our approach represents a promising step forward.

\subsection{Ablation Study for Distillation}
As shown in Table \ref{tab:predictor_only}, we provide additional experiments using the predictor directly on multi-step models without distillation to demonstrate how distillation facilitates predictor construction. While the `w/o distillation' baseline yields acceptable performance, it exhibits noticeable degradation as inference steps are reduced. In contrast, DisCa benefits from the smoother trajectories induced by distillation, which enable more precise predictions.
\begin{table}[htbp]
\caption{Ablation study for predictor w/ and w/o distillation}\label{tab:predictor_only}
\vspace{-9pt}
\resizebox{.48\textwidth}{!}{
\begin{tabular}{l|cc|ccc}
\toprule
\bf Method     &\bf NFE$\downarrow$      &\bf Latency(s)$\downarrow$   &\bf LPIPS$\downarrow$   &\bf PSNR$\uparrow$      & \bf SSIM$\uparrow$     \\
\midrule
\bf Predictor w/o distillation           & 8     & 145.5     & 0.1904      & 20.03      & 0.7278    \\
\bf Restricted MeanFlow(R=0.2) (Ours)    & 8     & 124.5     & 0.1778      & 19.85      & 0.7186    \\
\bf Predictor w/o distillation           & 4     & 90.4      & 0.2498      & 18.55      & 0.6963    \\
\bf {DisCa ($N=2, R=0.2$) (Ours)}                       & {4}   & {64.2}    & {0.1942}    & {19.29}    &{0.7071}   \\

\bottomrule
\end{tabular}
}

\vspace{-12pt}
\end{table}

\section{Experiment details}

\subsection{Training and Distillation settings}
Starting with the HunyuanVideo with Classifier-Free-Guidance (CFG), we first completed CFG distillation using a learning rate of $lr=10^{-5}$. Specifically, instead of inferring the CFG and No-CFG branches separately, we utilized a small FFN to append the input CFG information to the condition vector, thereby learning the behavior under different CFG settings. During this distillation process, the CFG was randomly sampled within the range of 1.0 to 8.0.

The model, after CFG distillation, already achieves a theoretical $2\times$ speedup compared to the original. Building on this, we proceeded with MeanFlow distillation (which is the Restricted MeanFlow discussed in this paper) using a learning rate of $lr=10^{-5}$. We conducted experiments with Restricted MeanFlow at $\mathcal{R}=0.4$ and $\mathcal{R}=0.2$. Given that the $\mathcal{R}=0.2$ version exhibited fewer artifacts, it was adopted as the foundation for DisCa training.

For DisCa training, we first used MSE loss (with a learning rate of $10^{-4}$) and random cache reuse intervals $\Delta = (t-t')$ sampled between 0 to $\Delta_{max} =0.2$ for a 500-iter initialization. We then introduced the discriminator for generative-adversarial training. The learning rate for the predictor was set to $10^{-4}$ and the discriminator's learning rate was set to $10^{-2}$, with the weight of generative-adversarial loss for predictor training $\lambda = 1.0$. Experiments confirmed that these settings ensured a stable adversarial dynamic between the two. The final results were obtained after 1000 iters of GAN training.

\subsection{Discussions on VBench}
VBench~\cite{VBench} is applied for evaluating the generated videos. VBench consists of 16 sub-dimensions that comprehensively assess video generation quality from multiple aspects. Among these, object class, multiple objects, human action, color, spatial relationship, scene, appearance style, temporal style, and overall consistency constitute the Semantic Score, measuring the model's semantic control capability. Meanwhile, subject consistency, background consistency, temporal flickering, motion smoothness, aesthetic quality, imaging quality, and dynamic degree form the Quality Score, evaluating the overall quality of the generated videos. These two scores are then weighted to produce the Total Score with a ratio of ``$\mathbf{Semantic: Quality = 1 : 4}$". 

Such a scoring strategy may partially reflect a model's capabilities and the knowledge it encapsulates. However, for critical issues in real-world application scenarios, such as \textbf{malformation, blurring, and other fatal flaws}, the Quality score, despite being weighted heavily, fails to respond effectively to them. This can be observed by comparing the visuals in Figure \ref{fig:Restricted_MFL_vis} and Figure \ref{fig:Vis-Huanyuan} with the metrics in Table \ref{table:Restriced Meanflow} and Table \ref{table:All Metrics}:
For instance, in Figure \ref{fig:Restricted_MFL_vis}, the MeanFlow 10-step example exhibits obvious malformation, while the Restricted MeanFlow 10-step example is free of such issues, yet the difference in their Quality scores is only 0.9\%. Besides, in Figure \ref{fig:Vis-Huanyuan}, the PAB output is completely blurred or even collapsed due to excessive acceleration, but the corresponding Quality score in Table \ref{table:All Metrics} only drops by 6.3\% compared to the non-accelerated model, suggesting relying on the Quality score or even the Total score as the primary metric for this paper is clearly unreasonable. 

Conversely, the \textbf{Semantic score responds well to such fatal issues}, frequently showing the most dramatic decline in the aforementioned scenarios characterized by distinct distortion and quality loss. Therefore, we elect to use the Semantic score as the primary metric, with the Quality and Total scores serving only as supplementary references.

\subsection{Break-down Analysis for VBench}
\begin{table*}[htbp]
\caption{\textbf{Detailed break-down analysis} for Semantic Score in Table \ref{table:All Metrics}.}\label{tab:vbench_breakdown}
\vspace{-10pt}
\resizebox{.98\textwidth}{!}{
\begin{tabular}{l|cccccccccc}
\toprule
\textbf{Method}                 &\bf Semantic &\bf Object &\bf Multiple &\bf Human  &\multirow{2}{*}{\bf Color}  &\bf Spatial    & \multirow{2}{*}{\bf Scene} & \bf Appearance    &\bf Temporal & \bf Overall\\
\textbf{HunyuanVideo 1.0 [T2V]}  & \bf Score   &\bf Class   &\bf Objects  &\bf Action  &                        &\bf Relationship&                        &\bf Style          &\bf Style     &\bf Consistency \\
\midrule
\textbf{Original:50steps}       &73.5\textcolor{gray!70}{\scriptsize (-0.0\%)} &85.4 \textcolor{gray!70}{\scriptsize (-0.0\%)}  &59.5\textcolor{gray!70}{\scriptsize (-0.0\%)}     &96.0 \textcolor{gray!70}{\scriptsize (-0.0\%)}    &85.3\textcolor{gray!70}{\scriptsize (-0.0\%)}            &58.5\textcolor{gray!70}{\scriptsize (-0.0\%)} & 46.2\textcolor{gray!70}{\scriptsize (-0.0\%)} &22.1\textcolor{gray!70}{\scriptsize (-0.0\%)}          &24.7\textcolor{gray!70}{\scriptsize (-0.0\%)}     &27.6\textcolor{gray!70}{\scriptsize (-0.0\%)} \\
\textbf{DisCa($R=0.2,N=4$)}     &69.3\textcolor{gray!70}{\scriptsize (-5.7\%)} &79.1\textcolor{gray!70}{\scriptsize (-7.4\%)}   &56.4\textcolor{gray!70}{\scriptsize (-5.2\%)}     &84.8\textcolor{gray!70}{\scriptsize (-11.7\%)}     &84.8\textcolor{gray!70}{\scriptsize (-0.6\%)}            &72.8\textcolor{gray!70}{\scriptsize (+24.4\%)} & 33.2\textcolor{gray!70}{\scriptsize (-28.1\%)} &20.8\textcolor{gray!70}{\scriptsize (-5.9\%)}          &22.9\textcolor{gray!70}{\scriptsize (-7.3\%)}     &26.3\textcolor{gray!70}{\scriptsize (-4.7\%)} \\
\bottomrule
\end{tabular}
}
\vspace{-6pt}
\end{table*}
We decomposed the variations in VBench Semantic Scores following acceleration. Analysis reveals that while scene-related performance decreased by $28.1\%$, Spatial Relationship scores improved by $24.4\%$, demonstrating superior spatial awareness. We leave the mitigation of these scene-related limitations for future exploration.
\begin{table*}[htbp]
\centering
\caption{\textbf{Comparison for the theoretical and actual acceleration of different methods} on HunyuanVideo.
}
\vspace{-3mm}
\setlength\tabcolsep{5.0pt} 
  \small
  \resizebox{0.98\textwidth}{!}{
  \begin{tabular}{l | c | c | c  c  c | c  c }
    \toprule
    {\bf Method} &{\bf CFG} & {\bf Peak} &\multicolumn{3}{c|}{\bf Theoretical} &\multicolumn{2}{c}{\bf Actual} \\
    \cline{4-6}
    \cline{7-8}
    {\bf HunyuanVideo 1.0 [T2V]\citep{sun_hunyuan-large_2024}} & {\bf Distilled} &{\bf VRAM}  & {\bf FLOPs(T) $\downarrow$} & {\bf Speed $\uparrow$}    &{\bf Latency(s)$\downarrow$} & {\bf Latency(s)  $\downarrow$}   & \bf Speed$\uparrow$  \\
    \midrule
  
  $\textbf{Original: 50 steps}$ 
                            & \ding{56}  & {99.23GB}  &394552.32 &{1.00$\times$}  &{1155.3} & {1155.3}  & {1.00$\times$}       \\ 

  $\textbf{CFG Distilled: 50 steps}$ 
                            & \ding{52}  & {97.21GB}  &197276.16 &{2.00$\times$}  &577.7  &  {581.1}  & {1.99$\times$}       \\ 
\midrule
  $\textbf{Original: 10 steps}$ 
                            & \ding{56}  & {99.23GB}  &78910.46 &{5.00$\times$}  &231.1  & {234.7}  & {4.92$\times$}        \\ 
  $\textbf{CFG Distilled: 20 steps}$ 
                             & \ding{52} & {97.21GB} &78910.46 &{5.00$\times$}  &231.1  &  {234.4}  & {4.93$\times$}      \\ 
  $\textbf{$\Delta$-DiT$(N=5)$}$ ~\cite{chen2024delta-dit}
                             & \ding{52}& {97.68GB}  &92068.57 &4.29$\times$ &269.3 &  {306.7}  & {3.77$\times$}       \\  
  $\textbf{PAB$(N=5)$}$ ~\cite{zhao2024PAB}
                             & \ding{52}  & {121.3GB} &64688.07 &6.10$\times$ &189.4 &  {216.5}  & {5.34$\times$}     \\  
  $\textbf{TeaCache$(l=0.15)$}$ ~\cite{liu2024timestep}
                            & \ding{52}& {97.70GB}  &75350.10 &5.24$\times$ &220.5 &  {237.6}  & {4.86$\times$}       \\
  $\textbf{FORA$(N=3)$}$ ~\cite{selvaraju2024fora}
                            & \ding{52} & {124.6GB} &78910.46 &{5.00$\times$}  &231.1 & {265.7}  & {4.35$\times$}       \\    
  $\textbf{TaylorSeer$(N=3,O=1)$}$ ~\cite{TaylorSeer2025}
                             & \ding{52}  & {130.7GB} &78910.46 &{5.00$\times$}  &231.1 &  {268.3}  & {4.31$\times$}       \\  
  $\textbf{MeanFlow: 20 steps}$ ~\cite{Geng2025MeanFF}
                             & \ding{52}& {97.21GB}  &78910.46 &{5.00$\times$}  &231.1 &  {232.7}  & {4.96$\times$}     \\ 
  $\textbf{Restricted MeanFlow: 20 steps}$[Ours]
                            & \ding{52}& {97.21GB}   &78910.46 &{5.00$\times$}  &231.1 &  {232.4}  & {4.97$\times$}      \\ 
\rowcolor{gray!20}
  $\textbf{DisCa$(\mathcal{R}=0.2, N=2)$}$ [Ours]
                            & \ding{52}& {97.64GB} &52239.36 &7.55$\times$ &153.0  & {152.8}  & {7.56$\times$}       \\ 
\midrule
  $\textbf{CFG Distilled: 10 steps}$ 
                            & \ding{52}& {97.21GB}  &39455.23 &{10.0$\times$} &115.5 &  {119.7}  & {9.65$\times$}      \\ 
  $\textbf{$\Delta$-DiT$(N=8)$}$ ~\cite{chen2024delta-dit}
                            & \ding{52} & {97.68GB}  &84178.00 &4.69$\times$ &246.3  &  {253.7}  & {4.55$\times$}      \\  

  $\textbf{PAB$(N=8)$}$ ~\cite{zhao2024PAB}
                            & \ding{52} & {121.3GB} &58058.68 &6.80$\times$ &169.9 &  {178.8}  & {6.46$\times$}       \\  
  $\textbf{TeaCache$(l=0.4)$}$ ~\cite{liu2024timestep}
                            & \ding{52}& {97.70GB} &40779.21 &9.68$\times$ &119.3 &  {125.3}  & {9.22$\times$}        \\  
  $\textbf{FORA$(N=6)$}$ ~\cite{selvaraju2024fora}
                            & \ding{52} & {124.6GB}&35509.71 &11.1$\times$ &108.3 &  {144.2}  & {8.01$\times$}       \\        
  $\textbf{TaylorSeer$(N=6,O=1)$}$ ~\cite{TaylorSeer2025}
                           & \ding{52} & {130.7GB} &43400.76 &9.09$\times$ &127.1 &  {166.0}  & {6.96$\times$}       \\  
  $\textbf{Restricted MeanFlow: 9 steps}$[Ours]
                          & \ding{52}& {97.21GB} &35509.71 &11.1$\times$ &104.0 & {108.3}  & {10.7$\times$}      \\ 

\rowcolor{gray!20}
  $\textbf{DisCa$(\mathcal{R}=0.2, N=3)$}$ [Ours]
                           & \ding{52}& {97.64GB}  &44619.05 &8.84$\times$ &130.7 &  {130.7}  & {8.84$\times$}       \\ 
\rowcolor{gray!20}
  $\textbf{DisCa$(\mathcal{R}=0.2, N=4)$}$ [Ours]
                            & \ding{52} & {97.64GB} &33188.56 &11.9$\times$ &97.1 &  {97.7}  & {{11.8}$\times$}      \\ 
    \bottomrule
  \end{tabular}}
  
  \label{table: acceleration_analysis}
\vspace{-5mm}
\end{table*}

\subsection{Comparison Configurations}
The acceleration methods mentioned above in Table \ref{table:All Metrics} can be categorized into the low-speed region (middle) and the high-speed region (lower). We briefly introduce their computational distribution configurations here:

\noindent\textbf{CFG Distilled \& (Restricted) MeanFlow:} Simply samples with the CFG-distilled model at a specified number of steps, such as 20 and 10 steps. The acceleration ratio is $2\times$ that of the original model (at the same given step count). 

\noindent\textbf{$\Delta$-DiT:} For the Multi-Modal DiT (MMDiT) structure in HunyuanVideo composed of 20 Double-Stream layers and 40 Single-Stream layers, $\Delta$-DiT skips the 40 Single-Stream layers via a residual-form cache at every cache step. After one complete computation step, it skips $(N-1)$ subsequent steps with cache. The configurations for the low- and high-speed regions are set to $N=5$ and $N=8$, respectively.

\noindent\textbf{PAB:} In the MMDiT architecture, the traditional distinction between Spatial-Temporal Attention and Cross-Attention does not exist. Therefore, the conventional hierarchical Attention caching strategy degenerates into periodically caching and reusing all Attention outputs. Here, the configurations for the low-speed and high-speed regions are set to $N=5$ and $N=8$, respectively.

\noindent\textbf{TeaCache:} TeaCache utilizes timestep embedding information to achieve dynamic adjustment of computational allocation. It  skips the cache step by performing \textbf{one} final reuse at the end. Here, the error threshold $\ell$ is set to $0.15$ for the low-speed region and $0.4$ for the high-speed region.

\noindent\textbf{FORA:} FORA achieves acceleration by caching and reusing the neural network parts within the residual networks corresponding to Attention and MLP in each DiT Block. Here, the configurations for the low-speed and high-speed regions are set to $N=3$ and $N=6$, respectively.

\noindent\textbf{TaylorSeer:} TaylorSeer's cache design is similar to FORA's but introduces caching with more derivative tensors, leading to higher VRAM occupancy but noticeably improved performance. Here, we align with the 3-step warm-up strategy from the original paper's code to protect structural information. The configurations for the low- and high-speed regions are $N=3$ and $N=6$, respectively.

\noindent\textbf{DisCa:} As introduced previously, DisCa performs further acceleration on the Restricted MeanFlow distilled 20-step model with the introduction of a lightweight learnable predictor. $N$ represents the maximum allowed cache interval during inference. For $N=4,3,2$, the inference alternates between 8,11,13 DiT inferences and corresponding 12,9,7 predictor inferences, respectively. Crucially, the inference cost of the predictor in these scenarios is almost negligible.

\begin{figure}
\centering\includegraphics[width=\linewidth]{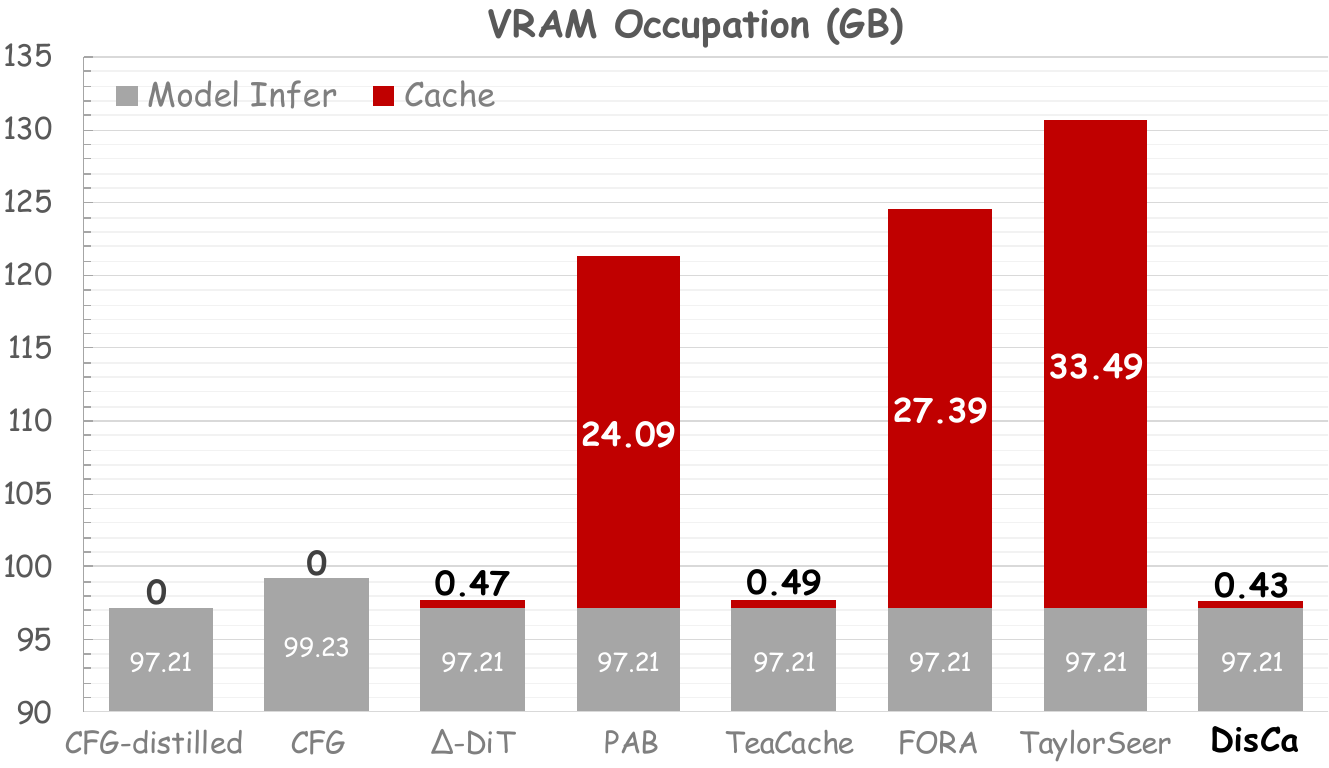}
\vspace{-7mm}
\caption{\textbf{VRAM Occupation analysis for different acceleration methods.} The VRAM consumption among different caching methods varies significantly; the proposed \textbf{DisCa} incurs only about $0.4 \text{ GB}$ of extra VRAM overhead, which is negligible.
}
\vspace{-6mm}

\label{fig:VRAM_Analysis}
\end{figure}

\subsection{Acceleration with Distributed Parallel}
\subsubsection{VRAM analysis}
As shown in Figure \ref{fig:VRAM_Analysis}, we analyzed VRAM consumption for each method, separating it into two parts: model with forward inference and cache occupancy. It can be observed that TeaCache, $\Delta$-DiT, and the proposed \textbf{DisCa} all incur negligible extra VRAM overhead. Conversely, caching schemes such as PAB, FORA, and TaylorSeer, which rely on multi-layer caching to provide richer information, generate significant VRAM overhead, increasing the demands on the operating device. Specifically, TaylorSeer, which performed optimally among previous methods, consumes an additional $33.49 \text{ GB}$ of VRAM. In contrast, \textbf{DisCa}, while achieving higher acceleration and higher quality (as shown in Table \ref{table:All Metrics}), incurs only $0.43 \text{GB}$ of additional VRAM overhead, compressing the extra VRAM cost to just $1.3\%$.

\begin{figure*}[h!]
\centering
\resizebox{0.98\textwidth}{!}{
\includegraphics[width=\linewidth]{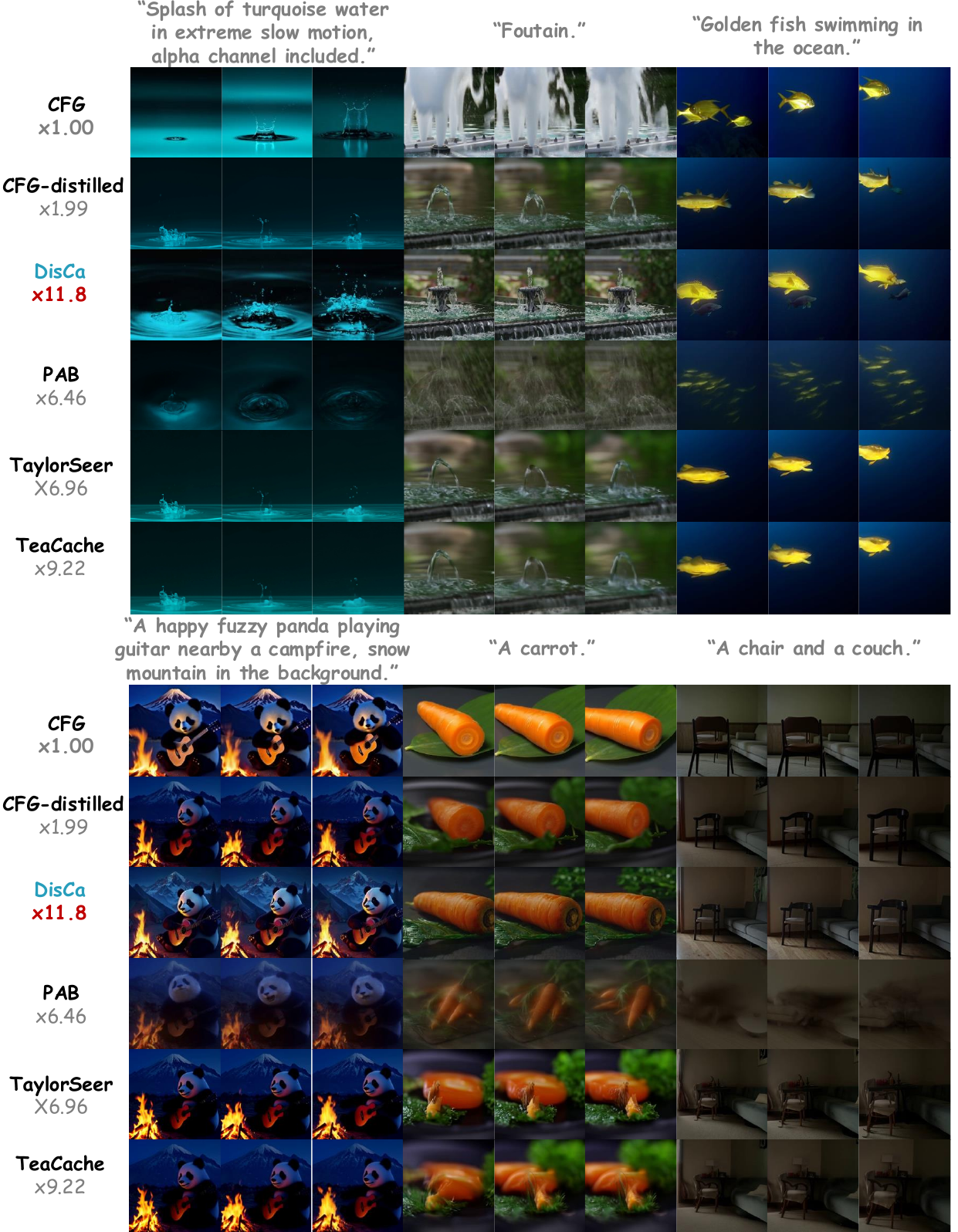}}
\vspace{-2mm}
\caption{\textbf{More visualization examples for HunyuanVideo.} The numerous examples clearly demonstrate that the proposed \textbf{DisCa} not only significantly surpasses previous acceleration schemes in terms of speed but also achieves \textit{immense advantages across multiple criteria, including structural semantics, detail fidelity, temporal consistency, adherence to physical plausibility, and aesthetic quality}.
}
\vspace{-7mm}
\label{fig:Vis-Huanyuan-appendix}
\vspace{-6mm}
\end{figure*}

\subsubsection{Cache Architecture Effects Actual Latency}
Prior works on feature caching have largely overlooked the relationship between the caching architecture and the discrepancy between theoretical and practical speedups. In this work, we analyze this gap and discuss its implications under the Distributed Parallel conditions.

As shown in Table \ref{table: acceleration_analysis}, we calculated the corresponding floating point operations (FLOPs) for each acceleration method and estimated their theoretical speedup ratio by referencing the FLOPs compression, which is then compared against the measured speedup ratio in Table \ref{table:All Metrics}.

As in Table \ref{table: acceleration_analysis}, the theoretical and actual speedup ratios for CFG Distilled, MeanFlow, and Restricted MeanFlow are extremely close. Among the cache acceleration methods, the theoretical and actual speedup ratios for TeaCache and $\Delta$-DiT are also well aligned. However, schemes such as PAB, FORA, and TaylorSeer exhibit a significant difference between their theoretical and actual speedup ratios. The closeness of the theoretical and actual speedup ratios for CFG Distilled, MeanFlow, and Restricted MeanFlow is obviously reasonable because they can be understood as simply reducing the number of computation steps, while the discrepancy observed among the cache acceleration methods, however, is worth considering. 

Those with a significant difference between theoretical and actual speedup in Table \ref{table: acceleration_analysis}, PAB, FORA, and TaylorSeer, all have a multi-layer cache structure. Specifically, they are multi-layer feature caching schemes that cache the output of each neural network layer individually, retaining the residual structure of the DiT Block to utilize richer features for computation. They prepare a set of cached tensors for every DiT Block: for PAB, it is the output of the attention calculation; for FORA, it is the output of both the attention and MLP calculations; and for TaylorSeer, it additionally includes their multi-order derivatives. 

As mentioned above, these schemes impose a particularly large VRAM overhead on the device. Furthermore, such a multi-layer cache structure generates frequent memory accesses during the cache step, and reusing these caches leads to multiple, albeit minor and especially sparse, calculations. In previous scenarios where sequence parallelism was not enabled, computational resources on the device were tight, and these memory accesses and sparse computations were optimized by low-level hardware libraries. Therefore, the extra overhead, though present, was not severe enough to warrant action. 

However, with the sequence parallelism technique enabled here, computational resources become relatively abundant. As a result, these sparse memory accesses and computations are not fully optimized by the low-level hardware libraries and the GPU device, leading to unexpectedly greater time consumption. In contrast, TeaCache and $\Delta$-DiT generate only one memory access and one simple, reused extra computation during a single cache-step inference, so their additional overhead is almost negligible.

\paragraph{Summary:} \textit{The cache structure is particularly critical in high-resolution, long-sequence video generation scenarios. It not only determines VRAM overhead but also significantly impacts computational efficiency in a parallel environment. Clearly, only reusing the cache in the final layer, instead of a complex multi-layer cache structure, is more appropriate. }
\vspace{2mm}

The proposed DisCa employs this single-final-layer-cache structure. Although the introduction of the additional small neural network predictor results in higher extra computation during the cache step than methods like TaylorSeer, which use simple calculations for prediction, this computational overhead remains negligible compared to a complete DiT inference. Furthermore, because the predictor calculation introduced by DisCa is inherently \textit{highly parallelized and imposes low memory access pressure}, it can achieve an acceleration ratio of up to $\mathbf{11.8\times}$, \textit{with the difference from the theoretical acceleration even in the margin of error}.

\subsection{More Visualization Results}
In this section, we provide additional visualization examples to further substantiate the superior improvement achieved by the proposed \textbf{DisCa} compared to previous acceleration schemes, as shown in Figure \ref{fig:Vis-Huanyuan-appendix}.


\section{Detailed Algorithms}

In this section, we present detailed algorithms as pseudo-code for the aforementioned CFG Distillation, Restricted MeanFlow Distillation, and Generative Adversarial Training process for the Predictor to facilitate understanding.

\begin{algorithm}[htbp]
    \textbf{Input:}  max and min cfg scale $g_{max},g_{min}$, 
    \\- \quad \quad~ data-noise-text pairs $\{x_0, \epsilon, c\}_i$,
    \\- \quad \quad~ CFG Model $\mathcal{M}^{CFG}$.\\
    \textbf{Init:} ~~~CFG Distilled Model $\mathcal{M}_{\theta}^* = \mathcal{M}^{CFG}$.
	\caption{{CFG distillation}}
    \label{alg:cfg_distill}
    \begin{algorithmic}[1] 
        
        \WHILE{Training}
        \STATE {Sample $t \sim U(0,1)$, $g \sim U(g_{min},g_{max})$ }.
        \STATE {Sample $x_t = (1-t)\cdot x_0 + t\cdot \epsilon$ }.
        \STATE Compute $v_{c} = \mathcal{M}^{CFG}(x_t, t, c)$.
        \STATE Compute $v_{uc} = \mathcal{M}^{CFG}(x_t, t, None)$.
        \STATE Compute $v_{target} = g \cdot v_{c} + (1-g) \cdot v_{uc}$.
        \STATE Compute $v_{\theta} = \mathcal{M}_{\theta}^*(x_t, t, c)$.
        \STATE Compute $\texttt{loss} = \|v_{\theta} - v_{target}\|_2^2$.
        \STATE \texttt{loss.backward(),optimizer.step()} 
        \\ \# \textit{Loss backward \& optimize parameter $\theta$.}
        \ENDWHILE
        \STATE \textbf{return} CFG Distilled Model $\mathcal{M}^*_{\theta}$. 
    \end{algorithmic}
\end{algorithm}

\begin{algorithm}[htbp]
    \textbf{Input:} data-noise pairs $\{x_0, \epsilon\}_i$,  
    \\- \quad \quad~ CFG distilled $\mathcal{M}^{*}$, 
    \\- \quad \quad~ Restrict factor $\mathcal{R}$. 
    \\- \quad \quad~ \# \textit{text info $c$ is omitted for simplification.} \\
    \textbf{Init:} ~~~Restricted MeanFlow model $\mathcal{M}_{\theta} = \mathcal{M}^*$
	\caption{{Restricted MeanFlow distillation}}
    \label{alg:restricted_meanflow}
    \begin{algorithmic}[1] 
        
        \WHILE{Training}
        \STATE {Sample $t, r = $\texttt{sample\_t\_r(}$\mathcal{R}$\texttt{)}}:
        \\ ~~~ \# ~~\textit{Sampling in Restricted MeanFlow.}
        \\ ~~~ \# \quad Sample $\mathcal{I} \sim U(0,\mathcal{R})$.
        \\ ~~~ \# \quad Sample $t \sim U(0,1)$.
        \\ ~~~ \# \quad Compute $r = \max(0, t-\mathcal{I})$.
        
        \STATE {Sample $x_t = (1-t)\cdot x_0 + t\cdot \epsilon$ }.
        \STATE {Compute $v = \mathcal{M}^{*}(x_t, t)$}.
        \STATE {Compute $u, \mathrm{d}u/\mathrm{d}t=$\texttt{jvp(}$\mathcal{M}_{\theta}, (x_t,r,t),(v,0,1)$\texttt{)}}
        \STATE Compute $u_{tgt} = v - (t-r)\cdot \mathrm{d}u/\mathrm{d}t$.
        \STATE Compute $\texttt{loss} = \|u - \texttt{stopgrad}(u_{tgt})\|_2^2$.
        \\ \# \textit{No gradient backward go through $u_{tgt}$.}
        \STATE \texttt{loss.backward(),optimizer.step()} 
        \\ \# \textit{Loss backward \& optimize parameter $\theta$.}
        \ENDWHILE
        \STATE \textbf{return} Restricted MeanFlow Distilled Model $\mathcal{M}_{\theta}$. 
    \end{algorithmic}
\end{algorithm}

\begin{algorithm}[htbp]
    \textbf{Input:} data-noise pairs $\{x_0, \epsilon\}_i$, 
    \\- \quad \quad~ Sampling timestep pairs $\{(t,r)\}_i$, 
    \\- \quad \quad~ Restricted MeanFlow distilled model $\mathcal{M}$, 
    \\- \quad \quad~ max cache interval  $\Delta_{max}$. 
    \\- \quad \quad~ Features Extractor $\mathcal{F}:=\mathcal{M}$,

    \textbf{Init:} ~~~random initialized Predictor $\mathcal{P}_{\theta_P}$, 
    \\- \quad \quad~ random initialized Discriminator $D_{\theta_D}$.
	\caption{{Predictor Training}}
    \label{alg:predictor_training}
    \begin{algorithmic}[1] 
        
        \WHILE{Training}
        \STATE {Sample $\Delta \sim U(0,\Delta_{max})$}.
        \STATE {Compute $(t',r')=\max((0,0),(t-\Delta, r-\Delta))$}.
        \STATE {Sample $x_t = (1-t)\cdot x_0 + t\cdot \epsilon$ }.
        \STATE {Compute $\mathcal{C}=\mathcal{M}(x_t,r,t)$}. ~~~~~\#~\textit{Initialize Cache}
        \STATE {Sample $x_{t'} = (1-t')\cdot x_0 + t'\cdot \epsilon$ }.
        \STATE {Compute $ u_{pred} = \mathcal{P}_{\theta_p}(\mathcal{C}, x_{t'}, r', t')$}.
        \STATE {Compute $ u_{tar} = \mathcal{M}(x_{t'}, r', t')$}.
        \STATE {Compute} $x_{t''}^{pred} = x_{t'} - (t'-r')\cdot u_{pred}$.
        \STATE {Compute} $x_{t''}^{tar} = x_{t'} - (t'-r')\cdot u_{tar}$.
        \STATE {Define} $(t'',r'') = (r', r'-(t''-r''))$.
        \STATE {Compute} $\mathcal{L}_{\mathcal{P}} = \|u_{pred} - u_{tar} \|_2^2 + \lambda \cdot \max(0, 1 - \mathcal{D}\circ\mathcal{F}(x_{t''}^{pred})),$
        \STATE {$\mathcal{L}_{\mathcal{P}}$\texttt{.backward(), optimizer\_P.step()}}
        \STATE {Compute} $\mathcal{L}_{\mathcal{D}}=\max(0, 1 - \mathcal{D}\circ\mathcal{F}(x_{t''}^{pred}))+\max(0, 1 + \mathcal{D}\circ\mathcal{F}(x_{t''}^{tar}))$.
        \STATE {$\mathcal{L}_{\mathcal{D}}$\texttt{.backward(), optimizer\_D.step()}}
       \\ \# \textit{Loss backward \& optimize parameter $\theta_P,\theta_D$.}
        \ENDWHILE
        \STATE \textbf{return} Restricted MeanFlow Distilled Model $\mathcal{M}_{\theta}$. 
    \end{algorithmic}
\end{algorithm}

\end{document}